\documentclass[10pt,twocolumn,letterpaper]{article}

\usepackage{cvpr}              %

\usepackage[utf8]{inputenc} %
\usepackage[T1]{fontenc}    %
\usepackage{url}            %
\usepackage{booktabs}       %
\usepackage{amsfonts}       %
\usepackage{nicefrac}       %
\usepackage{microtype}      %
\usepackage{xcolor}         %

\usepackage{graphicx}
\usepackage{amsmath}
\usepackage{amssymb}
\usepackage{booktabs}
\usepackage{pifont}%
\newcommand{\cmark}{\ding{51}}%
\usepackage{arydshln}
\usepackage{dblfloatfix}

\usepackage{colortbl} 
\usepackage{multirow} 
\newcommand*\rot{\rotatebox{90}}

\usepackage{caption}
\usepackage{wrapfig}

\usepackage{tikz}
\usetikzlibrary{bayesnet}
\usetikzlibrary{arrows}
\usepackage{float}

\usepackage{algorithm}
\usepackage[noend]{algpseudocode}
\usepackage{arydshln}

\usepackage[accsupp]{axessibility}
\usepackage{hyperref}       %

\usepackage[capitalize]{cleveref}
\crefname{section}{Sec.}{Secs.}
\Crefname{section}{Section}{Sections}
\Crefname{table}{Table}{Tables}
\crefname{table}{Tab.}{Tabs.}

\definecolor{cvprblue}{rgb}{0.21,0.49,0.74}

\def\paperID{8991} %
\def\confName{CVPR}
\def\confYear{2025}

\title{FALCON ~\includegraphics[height=20pt]{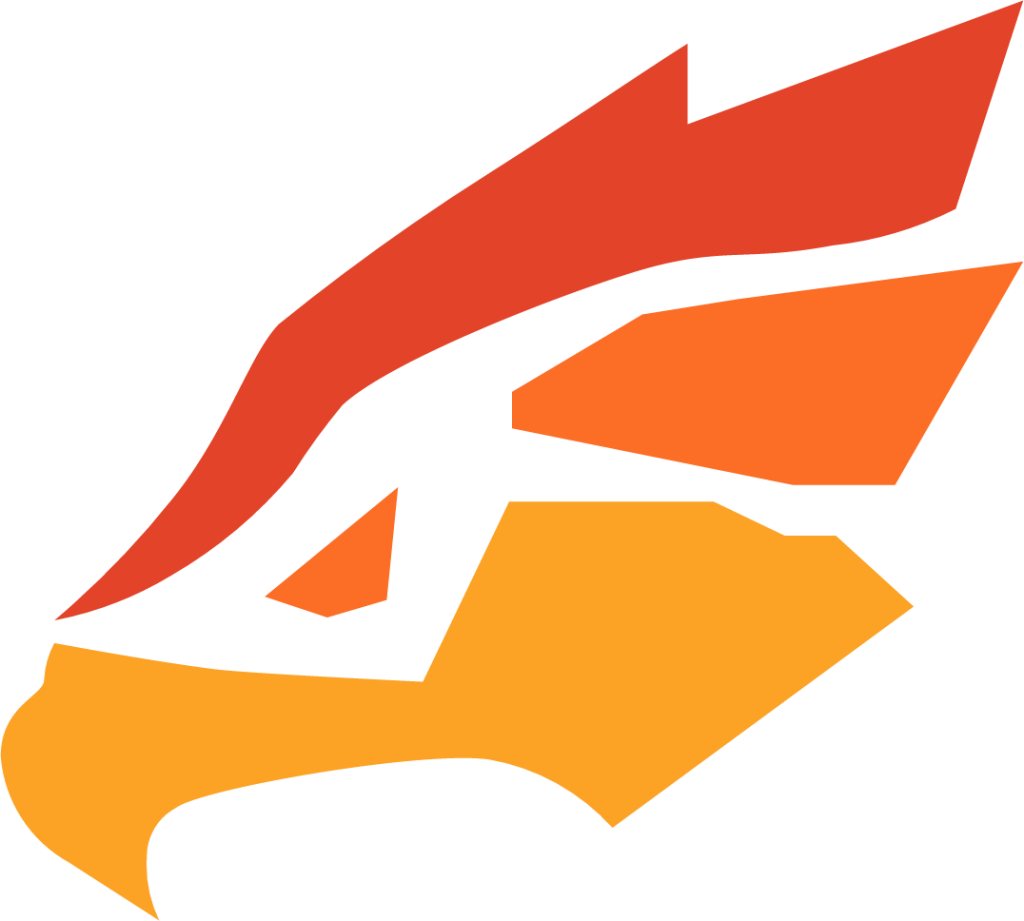}: Fairness Learning via Contrastive Attention Approach to\\Continual Semantic Scene Understanding
\vspace{-4mm}
}

\author{
Thanh-Dat Truong$^{1}$, Utsav Prabhu$^{2}$, Bhiksha Raj$^{3,4}$, Jackson Cothren$^{5}$, Khoa Luu$^{1}$\\
$^{1}$CVIU Lab, University of Arkansas, USA \quad
$^{2}$Google DeepMind, USA  \\
$^{3}$Carnegie Mellon University, USA  \quad
$^{4}$Mohammed bin Zayed University of AI, UAE\\
$^{5}$Dep. of  Geosciences, University of Arkansas, USA \\
\tt\small \{tt032, jcothre, khoaluu\}@uark.edu, bhiksha@cs.cmu.edu, utsavprabhu@google.com\\
\small{\url{http://uark-cviu.github.io/projects/FALCON}}
\vspace{-4mm}
}

\begin{document}
\maketitle

\begin{abstract}

Continual Learning in semantic scene segmentation aims to continually learn new unseen classes in dynamic environments while maintaining previously learned knowledge. Prior studies focused on modeling the catastrophic forgetting and background shift challenges in continual learning. However, fairness, another major challenge that causes unfair predictions leading to low performance among major and minor classes, still needs to be well addressed. In addition, prior methods have yet to model the unknown classes well, thus resulting in producing non-discriminative features among unknown classes. This work presents a novel Fairness Learning via Contrastive Attention Approach to continual learning in semantic scene understanding. In particular, we first introduce a new Fairness Contrastive Clustering loss to address the problems of catastrophic forgetting and fairness. Then, we propose an attention-based visual grammar approach to effectively model the background shift problem and unknown classes, producing better feature representations for different unknown classes. Through our experiments, our proposed approach achieves State-of-the-Art (SoTA) performance on different continual learning benchmarks, i.e., ADE20K, Cityscapes, and Pascal VOC. It promotes the fairness of the continual semantic segmentation model.
\end{abstract}

\section{Introduction}

\begin{figure}[!t]
    \centering
    \includegraphics[width=1.0\linewidth]{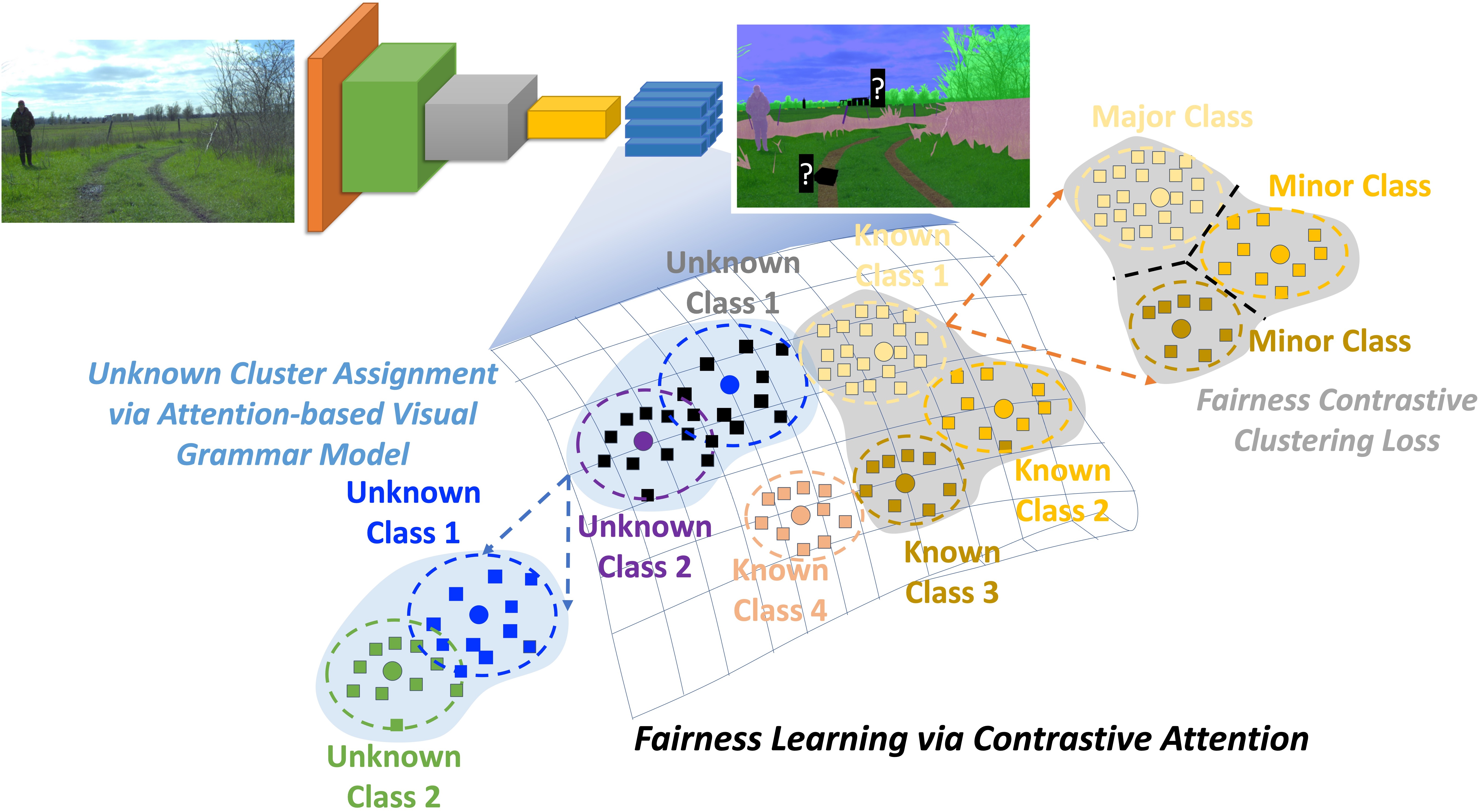}
    \vspace{-6mm}
    \caption{\textbf{Our Fairness Learning via Contrastive Attention to Continual Semantic Segmentation.} The \textit{Fairness Contrastive Clustering Loss} promotes the fairness of the model while the \textit{Attention-based Visual Grammar} models the unknown classes. 
    } \label{fig:highlight}
    \vspace{-4mm}
\end{figure}

The semantic segmentation networks, e.g., Transformers \cite{xie2021segformer} and Convolutional Neural Networks \cite{chen2018deeplab}, 
learned from data with a closed-set of known classes, have shown outstanding performance.
However, they often suffer performance degradation when encountering novel objects or classes in new dynamic environments \cite{douillard2021plop, cermelli2020modelingthebackground, truong2023fairness, truong2024conda}. To improve their performance, several transfer learning and domain adaptation methods \cite{daformer, Araslanov:2021:DASAC, truong2023fairness, truong2021bimal, truong2024eagle, truong2025cross, nguyen2022self, truong2022otadapt, jalata2022eqadap, truong2020domain} were introduced to adapt trained models into deployed environments. While the former often aims to fine-tune the model on labeled data collected in the new environments, the latter adapts the model to the new domains in an unsupervised manner \cite{Araslanov:2021:DASAC, daformer}. However, these methods cannot handle novel objects well due to their close-set learning.
In practice, the semantic segmentation models should be able to adaptively and continually learn the new knowledge of novel classes. It motivates the development of \textbf{\textit{Continual Learning}} paradigm \cite{cermelli2020modelingthebackground, douillard2021plop, sats_prj_2023}, a.k.a, \textbf{\textit{Continual Semantic Segmentation}} (CSS), where the segmentation models are learned sequentially to new contents of data.

Far apart from prior segmentation methods \cite{chen2018deeplab, xie2021segformer} that learn one time on static, closed-set data, Continual Learning requires the segmentation models to learn from dynamic, open-set data \cite{douillard2021plop, truong2023fairness}. In a particular scenario, accessing previous learning data is restricted due to privacy concerns. In CSS, three challenges have been identified, including (1) Catastrophic Forgetting, (2) Background Shift, and (3) Fairness. 
While the catastrophic forgetting problem \cite{robins1995catastrophicforgetting,french1999catastrophicforgetting,thrun1998lifelonglearning} depicts the segmentation model tends to forget its knowledge when learning new data, 
background shift indicates the problem of classes of previous or future data (unknown classes) have collapsed into a background class \cite{cermelli2020modelingthebackground, douillard2021plop, zhang2022representation}. 
Prior methods \cite{douillard2021plop, sats_prj_2023, ssul_neurips_2021, truong2023fairness} addressed these two problems by introducing knowledge distillation and pseudo labels.
However, these methods can not handle unknown classes since they either consider these unknown classes as a background class or assign unknown pixels by a pseudo label of prior known classes \cite{douillard2021plop, truong2023fairness}.
More importantly, the last problem, \textit{fairness}, is a significant challenge that limits the performance of CSS models (Fig. \ref{fig:highlight}). 

\begin{figure}[!t]
    \centering
    \includegraphics[width=1.0\linewidth]{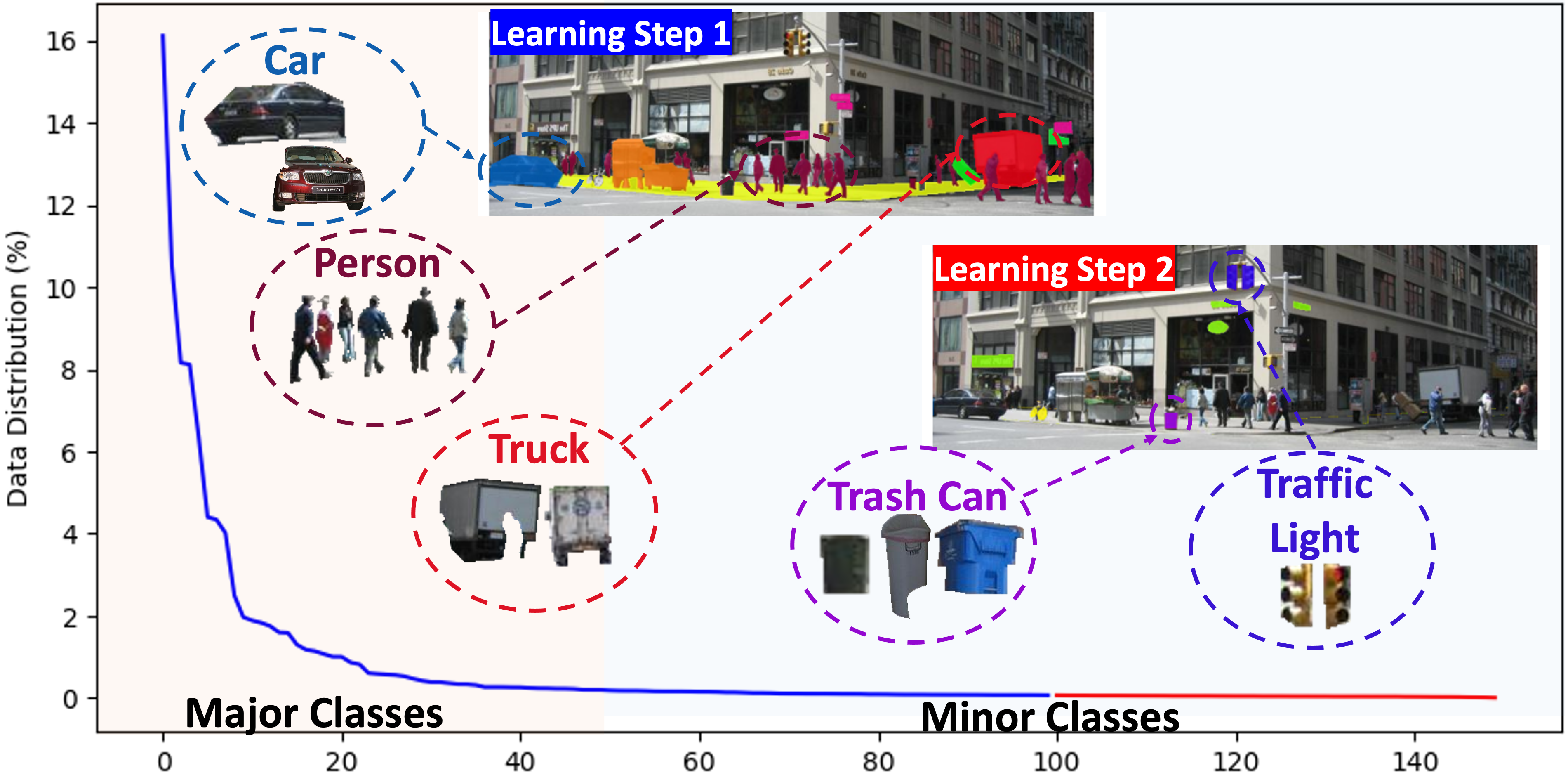}
    \vspace{-6mm}
    \caption{\textbf{The Data Class Distribution of ADE20K.} The major classes occupy more than 75\% of the total pixels of the dataset.}
    \label{fig:data_class_dist}
    \vspace{-6mm}
\end{figure}

As shown in Fig. \ref{fig:data_class_dist}, the \textit{number of pixels of each class} in training data have been imbalanced among classes and significantly decreased after each task. Thus, this bias influences the learning procedure and model predictions that later cause unfair class predictions. However, limited studies are taking the fairness problem into account. \cite{Truong:CVPR:2023FREDOM} presented a similar problem in domain adaptation and extended it to continual learning \cite{truong2023fairness}. These methods rely on the assumption of ideal fair or balanced data distributions. However, it is not applicable in practice since the size, i.e., the number of pixels, of several classes can never be more significant than others. For example, the bottle size should not be more significant than the size of a car.
Meanwhile, the current knowledge distillation methods \cite{douillard2021plop, cswkd_cvpr_2022, cermelli2023comformer} in CSS are unable to handle the fairness problem since they focus on modeling catastrophic forgetting and background shift problems.
Therefore, it is essential to develop a new CSS approach to address these limitations.

\noindent
\textbf{Contributions of This Work.} This work presents a novel {\textit{\textbf{Fa}irness \textbf{L}earning via \textbf{Con}trastive Attention Approach (\textbf{FALCON})}} to Continual Semantic Segmentation (as shown in Fig. \ref{fig:highlight}). 
Our contributions can be summarized as follows. 
First, we introduce a novel \textit{\textbf{Contrastive Clustering Paradigm}} approach to Continual Learning that models the catastrophic forgetting problem. Second, by analyzing the limitation of vanilla Contrastive Clustering in biased data, we introduce a novel \textit{\textbf{Fairness Contrastive Clustering}} loss to model the fairness problem in continual learning efficiently. Third, to effectively model the background shift problem, we introduce a new \textit{\textbf{Attention-based Visual Grammar}} 
that model the topological structures of feature distribution to handle the unknown classes effectively.
Finally, the ablation studies illustrate the effectiveness of the proposed approach in different aspects of fairness promotion in CSS models. Compared with prior methods, our approach achieves SoTA performance on different settings of three standard benchmarks of CSS, including ADE20K, Pascal VOC, and Cityscapes.

\section{Related Work}

\textbf{Continual Semantic Segmentation.} 
Several studies were introduced to address catastrophic forgetting and background shift problems \cite{Ermis_2022_CVPR, kirkpatrick2017ewc, lopezpaz2017gem, douillard2020podnet, douillard2021plop, ozdemir2018learn, ozdemir2019extending, michieli2019incremental}.
The common CSS approach adopts knowledge distillation \cite{douillard2021plop} and pseudo labels \cite{cermelli2020modelingthebackground} to model catastrophic forgetting and background shift, respectively.
Later, it was further improved by decoupling knowledge representations \cite{zhang2022representation}, modeling the inter- and intra-class knowledge \cite{sats_prj_2023, cong2025cs},  distinguishing the feature representations of the future classes \cite{ssul_neurips_2021, cswkd_cvpr_2022, wang2024incremental, xie2025early}, reducing background confusion \cite{yang2023label, zhang2024background}, using visual prompting \cite{kim2024eclipse, liu2024learning}, 
or modeling distillation loss via the geodesic flow \cite{simon2021learning}.
Another approach \cite{cermelli2023comformer, gong2024continual, chen2024strike} adopted the mask-based segmentation networks \cite{cheng2021maskformer, cheng2022masked} to improve the performance.
Recent studies have introduced CSS under the unsupervised domain adaptation settings \cite{volpi2021continual, rostami2021lifelong, saporta2022muhdi, truong2024conda}.
However, prior studies have yet to well model different unknown classes. Particularly, as previous methods  \cite{douillard2021plop, cermelli2023comformer} use the pseudo labels to model the unknown classes, the future classes will be treated as a single background class. 
Then, Joseph et al. \cite{joseph2021towards}
improves the unknown class modeling using clustering but this method considers all different unknown classes as a single cluster, leading to the non-discriminative features among unknown classes.

\noindent
\textbf{Contrastive Learning (CL).} CL is a common learning approach \cite{oord2018representation, chen2020improved, chen2020simple} to structure the deep feature representations in the deep latent space.
Oorde et al. \cite{oord2018representation} first introduced the Noise-Contrastive Estimation (InfoNCE) learning framework. Then, Chen et al. \cite{chen2020simple} presented SimCLR, a self-supervised contrastive learning approach to improve the representation power of Residual Networks.
He \cite{he2020momentum} proposed a Momentum Contrast framework for unsupervised representation learning.
Later, it was further improved by using MLP projection head \cite{chen2020improved} and extended to improve the self-supervised training process of vision transformers \cite{chen2021mocov3}.
Cui et al. \cite{cui2021parametric} introduced a supervised parametric contrastive learning loss to address the long-tailed recognition. 
Li et al. \cite{li2021contrastive} adopted contrastive learning to develop the one-stage online contrastive clustering method. Radford et al. \cite{radford2021learning} presents a contrastive framework to learn the vision-language model.
Later, several methods also adopted this framework for vision-language pretraining \cite{radford2021learning, lei2021less}.

\noindent
\textbf{Imbalanced and Fairness Learning.}
The early methods utilized the balanced Softmax loss \cite{ren2020balanced} to alleviate the impact of imbalanced data distribution.
Later, Wang et al.  \cite{wang2021seesaw} introduced a Seesaw loss to re-balance the contributions of positive and negative instances via the mitigation and compensation modules. 
Ziwei et al.  \cite{liu2019largescale} introduced a dynamic meta-embedding to model the imbalanced classification problem.
Chu et al.  \cite{chu2021learning} reduce the bias in the segmentation model by presenting a new stochastic training scheme.
Szabo et al.  \cite{szabo2021tilted} presented a tilted cross-entropy loss to promote class-relevant fairness.
However, there are limited studies that address the fairness problem in CSS.
Truong et al. \cite{Truong:CVPR:2023FREDOM} introduced a fairness domain adaptation approach to semantic segmentation and later extended it into continual learning setting \cite{truong2023fairness}.
However, these methods \cite{Truong:CVPR:2023FREDOM, truong2023fairness} rely on the assumption of ideal balanced data, which could not be achieved by nature.
To address the limitations in prior work, \textit{this paper will introduce a novel approach to effectively model the fairness problem and unknown classes in the continual learning setting}.

\begin{figure*}[!t]
    \centering
    \includegraphics[width=0.8\textwidth]{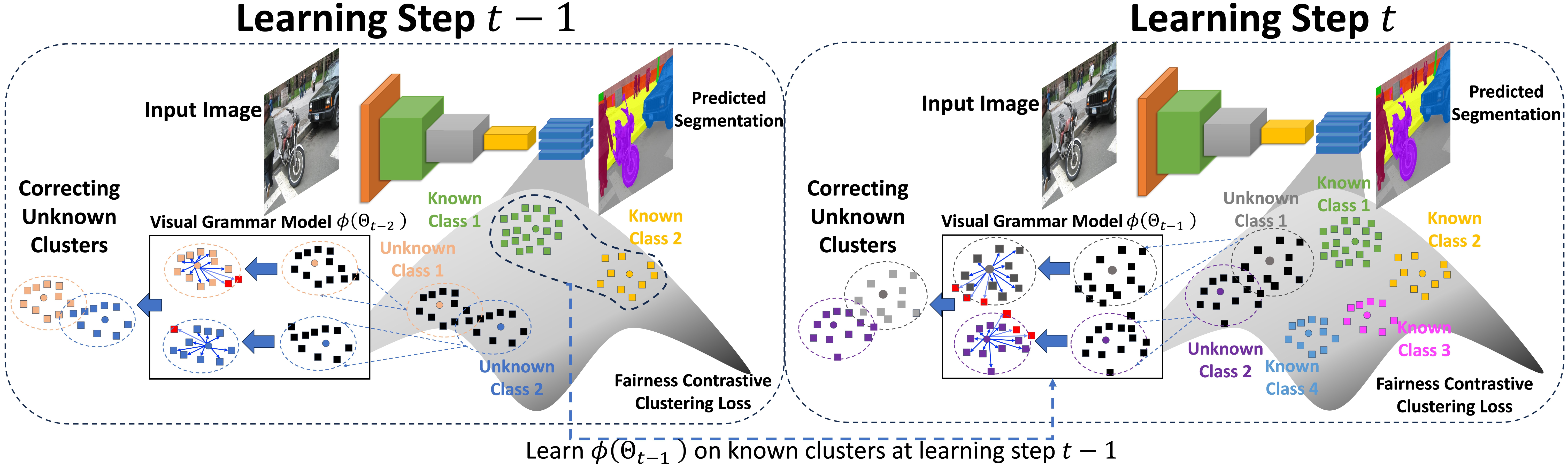}
    \vspace{-3.5mm}
    \caption{\textbf{The Proposed Fairness Learning via Contrastive Attention Approach to
Continual Semantic Scene Understanding.}}\label{fig:css_framework}
    \vspace{-5mm}
\end{figure*}

\section{The Proposed FALCON Approach}

CSS aims to learn a segmentation network $F$ on sequence data $\mathcal{D} = \{\mathcal{D}^1, ..., \mathcal{D}^T\}$ where $T$ is the number of learning steps. At learning step $t$, the model $F$
encounters a dataset $\mathcal{D}_t = \left\{(\mathbf{x}^t, \mathbf{\hat{y}}^t)\right\}$ where $\mathbf{x}^t \in \mathbb{R}^{H \times W \times 3}$ is the image and $\mathbf{y} \in \mathbb{R}^{H \times W}$ is a segmentation label of $\mathbf{x}^t$. 
The ground truths at learning step $t$ only consist of current classes $\mathcal{C}^t$, while the class labels of the previous $\mathcal{C}^{1...t-1}$ and future steps $\mathcal{C}^{t+1...T}$ are collapsed into a background class. 
Formally, learning the CSS model at step $t$ can be formed as Eqn. \eqref{eqn:general_obj}.
\begin{equation} 
\footnotesize
\label{eqn:general_obj}
\theta_t^* = \arg\min_{\theta_t} \mathbb{E}_{\mathbf{x}^t, \mathbf{\hat{y}}^t \in \mathcal{D}^t} \left[\mathcal{L}_{CE}\left(\mathbf{y}^t,  \mathbf{\hat{y}}^t\right) + \lambda_{CL}\mathcal{L}_{CL}\left(F(\mathbf{x}^t)\right)\right]
\end{equation}
where, $\mathbf{y}^t = F(\mathbf{x}^t, \theta_t)$, 
$\theta_t$ is the parameter of $F$ at current learning step $t$,
$\mathcal{L}_{CE}$ is the cross-entropy loss,
$\lambda_{CL}$ is the balanced weight.
and $\mathcal{L}_{CL}$ is the CSS objective.
At learning step $t$, the segmentation model $F$ is required to be able to predict both previously learned classes $\mathcal{C}^{1...t-1}$ and current new classes $\mathcal{C}^{t}$. 
Under this learning scenario, 
three challenges have been identified, i.e., Catastrophic Forgetting, Background Shift, and Fairness. Several prior methods were presented to model the two first issues in CSS using knowledge distillation \cite{douillard2021plop, cermelli2023comformer}. The last issue has not been well addressed yet due to its challenges \cite{truong2023fairness}. 
Prior methods \cite{shmelkov2017incremental, cermelli2020modelingthebackground, douillard2021plop, zhang2022representation, cermelli2023comformer} adopt knowledge distillation to design $\mathcal{L}_{CL}$.
However, this method prevents the CSS model from diverging knowledge learned previously, therefore resulting in limiting the ability to adopt new knowledge \cite{truong2023fairness}.
In addition, these methods have not addressed fairness and background shift problems due to their dedicated design for maintaining knowledge via distillation \cite{douillard2021plop, cermelli2023comformer, cswkd_cvpr_2022}. Therefore, to address these problems, we introduce a novel \textbf{\textit{Fairness Learning via Contrastive Attention Approach}} to CSS.

\subsection{Continual Learning via Contrastive Clustering}

Apart from prior methods \cite{douillard2021plop, cermelli2023comformer, cswkd_cvpr_2022}, our CSS (Fig. \ref{fig:css_framework}) is defined as Contrastive Clustering Learning.
Given a set of centroid vectors $\{\mathbf{c}_i\}_{i=1}^{N_K + N_U}$ where $N_K = |\mathcal{C}^{1..t}|$ and $N_U$ is the number of known and unknown classes up to current learning tasks.
Prior work \cite{joseph2021towards, truong2023fairness, cermelli2023comformer} often defined the number of unknown classes as $1$ where background classes are considered as a single unknown class. 
Formally, our Contrastive Clustering Learning for CSS can be defined as in Eqn. \eqref{eqn:general_clustering}.
\begin{equation} 
\footnotesize
\label{eqn:general_clustering}
\begin{split}
\mathcal{L}_{CL}\left(F(\mathbf{x}^t)\right) &= \sum_{\mathbf{c}_i}\mathcal{L}_{Cont}(\mathbf{F}^t, \mathbf{c}_i) \\
&=  \sum_{\mathbf{c}_i}\sum_{h, w}-\phi(\mathbf{f}^t_{h,w}, \mathbf{c}_i)\log \frac{\exp(\mathbf{f}^t_{h,w} \times \mathbf{c}_i)}{\sum_{\mathbf{f}'}\exp(\mathbf{f}' \times \mathbf{c}_i)}
\end{split}
\end{equation}
where $\mathbf{F}^t \in \mathbb{R}^{H \times W \times D}$ 
is the feature maps extracted from the input image $\mathbf{x}^t$ by the segmentation network $F$, 
$\mathbf{f}^t_{h,w} \in \mathbb{R}^{D}$ is the feature at the pixel location $(h , w)$ of features $\mathbf{F}^t$,
$\sum_{\mathbf{f}'}$ means the summation over all feature representations $\mathbf{f}' \in \mathbb{R}^{D}$, 
and $\phi: \mathbb{R}^{D} \times \mathbb{R}^{D}  \to [0, 1]$ is the function that determines either $\mathbf{f}^t_{h,w}$ belongs to the cluster $\mathbf{c}_i$ or not.

By defining CSS as contrastive clustering learning, the knowledge of the segmentation model has been well maintained via the cluster vectors $\mathbf{c}$.
Then, minimizing 
Eqn. \eqref{eqn:general_clustering} will separate the representations of different classes 
while gathering the features of the same class into the same cluster.
As the cluster vectors $\mathbf{c}$ of the old classes $\mathcal{C}^{1..t-1}$ have been well learned to represent their knowledge,
these vectors are frozen at learning step $t$ to maintain the knowledge representations of previous classes to address the catastrophic forgetting problem. 
To effectively learn cluster vectors $\mathbf{c}$, 
the cluster vector $\mathbf{c}$ will periodically updated after each $K$ steps by the momentum update \cite{joseph2021towards, truong2023fairness, he2020momentum} based on the features $\mathbf{f}^t_{h,w}$ assigned to cluster $\mathbf{c}$.
However, there are two major problems in contrastive clustering learning.
First, since the training data suffer the bias among classes as shown in Fig. \ref{fig:data_class_dist}, 
this bias will influence Eqn. \eqref{eqn:general_clustering} and cause the unfair predictions. 
Second, as the function $\phi$ requires the labels to determine the features belonging to clusters, it limits the ability to model the unknown classes where the labels are unavailable.
Therefore, Secs. \ref{sec:fairness}-\ref{sec:unknown_class_modeling}  will present a novel approach to tackle these problems.

\subsection{Fairness Contrastive Clustering Learning}
\label{sec:fairness}

While contrastive clustering learning in Eqn. \eqref{eqn:general_clustering} promotes the compact representations of features around their clusters, inspired by \cite{zhu2022balanced, cui2021parametric, chen2022perfectly}, we observe that the imbalanced class distribution will influence unfair behaviors among classes. 
In particular, for simplicity, we consider $\{\mathbf{f}^t_i\}_{i=1}^{L}$ is the set of features that belong to the cluster $\mathbf{c}$ at learning step $t$ (i.e., $\phi(\mathbf{f}_i^{t}, \mathbf{c}) = 1$) and $L$ is the number of features (in this case, $L$ is the total number of pixels belong to the class of cluster $\mathbf{c}$). Let us define the enforcement between the feature $\mathbf{f}^t_t$ and the cluster $\mathbf{c}$ as $\ell_i = \frac{\exp(\mathbf{f}^t_{i} \times \mathbf{c})}{\sum_{\mathbf{f}'}\exp(\mathbf{f}' \times \mathbf{c})}$. Hence, 
the lower the value of the enforcement $\ell_i$ is, the more compact the representation of visual features and clusters is.
Then, the contrastive clustering learning loss in Eqn. \eqref{eqn:general_clustering} of entire cluster $\mathbf{c}$ can be defined as in Eqn. \eqref{eqn:loss_for_one_cluster}.
\begin{equation}\label{eqn:loss_for_one_cluster}
\small
    \mathcal{L}_{Cont}(;, \mathbf{c}) = -\sum_{i=1}^L\log\frac{\exp(\mathbf{f}^t_{i} \times \mathbf{c})}{\sum_{\mathbf{f}'}\exp(\mathbf{f}' \times \mathbf{c})} = -\sum_{i=1}^L\log \ell_i 
\end{equation}

\noindent
\textbf{Proposition 1}. \textit{If the contrastive clustering loss $\mathcal{L}_{Cont}(;, \mathbf{c})$ achieves the optimal value, the enforcement $\ell_i$ between the feature and the cluster will converge to $\ell_i = L^{-1}$.}

\begin{figure}[!b]
    \centering
    \vspace{-4mm}
    \includegraphics[width=1.0\linewidth]{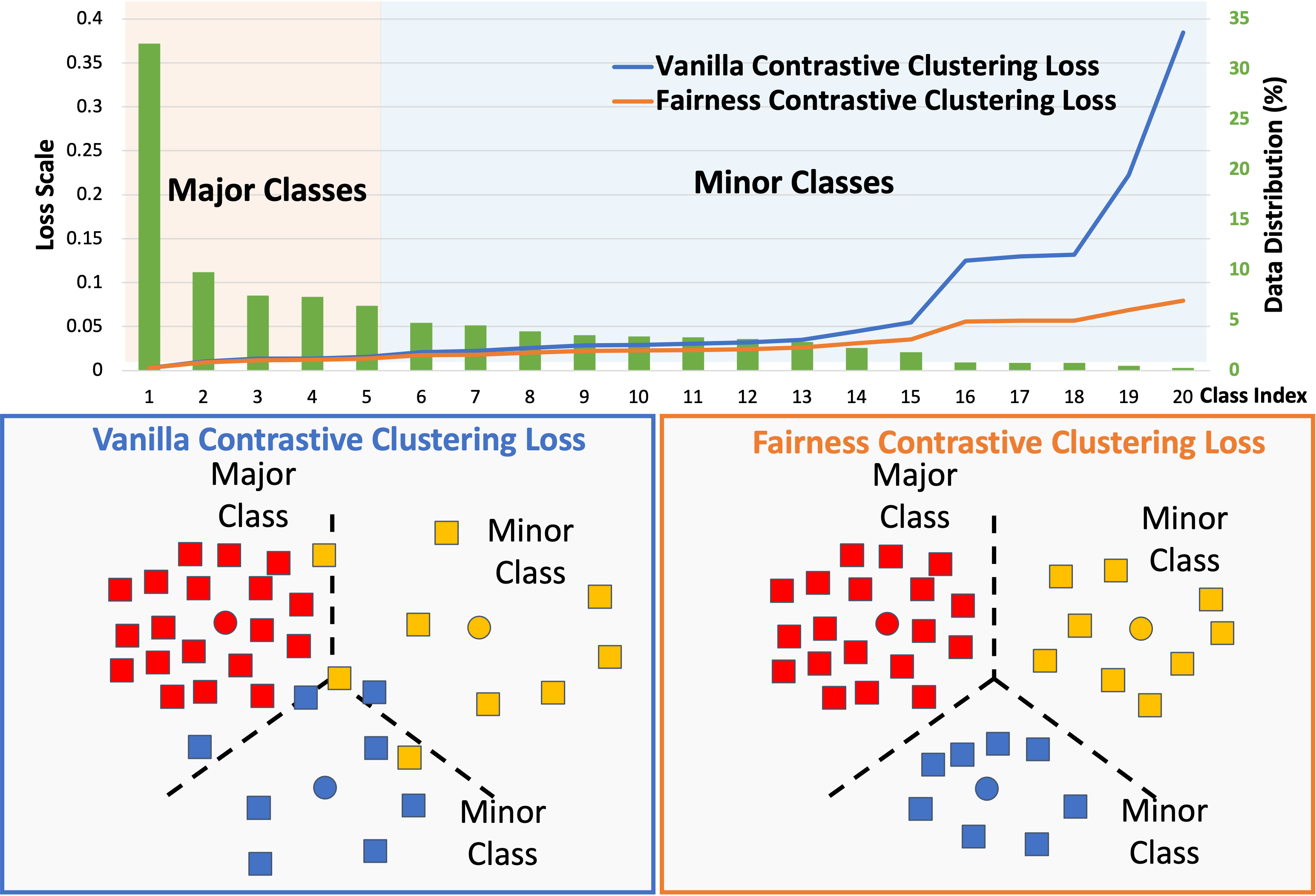} 
    \vspace{-6mm}
    \caption{\textbf{The Enforcement Loss of Contrastive Clustering $\mathcal{L}_{Cont}$ and Fairness Contrastive Clustering $\mathcal{L}^{\alpha}_{Cont}$ on Pascal VOC}.
    Since $\mathcal{L}_{Cont}$ suffers severe biased, its clusters of minor classes remain scattered.
    Our $\mathcal{L}^{\alpha}_{Cont}$ produces a more uniform loss among classes, promotes fairness and compactness of clusters.
    }
    \label{fig:impact_biased_data}
\end{figure}

\textbf{\textit{Proposition 1}} has implied that the class with more samples will result in a lower value of the enforcement and produce a more compact representation while the class having fewer samples will be more scattered in the feature space due to the higher value of the enforcement. In particular, let $L_{major}$ and $L_{minor}$ be the number of samples of the major and minor class where $L_{major} >> L_{minor}$. Then, based on \textbf{\textit{Proposition 1}}, the enforcement between features and the cluster of the major class will be significantly lower than the one of the minor class, i.e., $L_{major}^{-1} << L_{minor}^{-1}$.
Therefore, a direct adoption of the contrastive clustering loss in Eqn. \eqref{eqn:general_clustering} will result in an unfair CSS model.
In addition, for classes in the minority group, the lower value of $L$ results in the feature presentations of classes being far away from their clusters. Thus, the model will produce non-discriminative features compared to the ones in the majority group. Moreover, if the loss is applied to the cases of unknown labels, these feature representations can be scattered in the latent space and pulled into the incorrect clusters due to weak enforcement between features and clusters (Fig. \ref{fig:impact_biased_data}).

To address the unfair problem in contrastive clustering learning, 
inspired by \cite{zhu2022balanced, cui2021parametric, chen2022perfectly}, 
we introduce a scaling factor $\alpha$ and a learnable transition vector $\mathbf{v}$ for each cluster $\mathbf{c}$ (all clusters have the same value of $\alpha$ but different vector $\mathbf{v}$).
\textbf{\textit{ Our Fairness Contrastive Clustering Learning Loss}} for the entire cluster in Eqn. \eqref{eqn:loss_for_one_cluster} can be re-formed as follows,
\begin{equation}\label{eqn:loss_for_one_cluster_alpha}
\footnotesize
    \mathcal{L}^{\alpha}_{Cont}(;, \mathbf{c}) = -\alpha\sum_{i=1}^L\log\frac{\exp(\mathbf{f}^t_{i} \times \mathbf{c})}{\sum_{\mathbf{f}'}\exp(\mathbf{f}' \times \mathbf{c})} -\log\frac{\exp(\mathbf{v} \times \mathbf{c})}{\sum_{\mathbf{f}'}\exp(\mathbf{f}' \times \mathbf{c})}
\end{equation}
Intuitively, the scaling factor $\alpha$ will help to re-scale the impact of the enforcement in learning, and the transitive vector $\mathbf{v}$ assists in translating the center cluster into the proper position of the latent space. 
This action promotes the compactness of clusters in the minority group.

\noindent
\textbf{Proposition 2}. \textit{If the fairness contrastive clustering loss $\mathcal{L}^{\alpha}_{Cont}(;, \mathbf{c})$ achieves the optimal value, the enforcement $\ell_i$ between the feature and the cluster will converge to $\ell_i = (\alpha^{-1} + L)^{-1}$.} 
\textit{Proofs of Propositions 1-2 are in the appendix.}

Under the \textbf{\textit{Proposition 2}}, when the value of $\alpha$ is small, the divergence of the enforcement between major and minor classes will be smaller, i.e., $||(\alpha^{-1}+L_{major})^{-1} - (\alpha^{-1}+L_{minor})^{-1}|| < || L^{-1}_{major} - L^{-1}_{minor}|| $. Fig. \ref{fig:impact_biased_data} has illustrated the impact of fairness contrastive clustering loss. Therefore, our designed proposed fairness contrastive loss has effectively addressed the fairness issue in Eqn. \eqref{eqn:general_clustering}. 
It should be noted that although the smaller $\alpha$ results in the fairer enforcement varied from major to minor classes. However, if the value of scaling factor $\alpha$ is too small, the contrastive clustering loss will rely more on the enforcement of the transitive vector $\mathbf{v}$, and the distribution of features $\mathbf{f}^t_i$ around its cluster $\mathbf{c}$ will be scattered due the weak enforcement caused by small $\alpha$.
Therefore, the value of scaling factor $\alpha$ in practice should be carefully selected.

\subsection{An Efficient Unknown Class Modeling}
\label{sec:unknown_class_modeling}

An ideal CSS approach must be able to model the unknown classes without supervision, especially in open-set contexts \cite{joseph2021towards, truong2023fairness} where there could be multiple unknown classes or objects.
Prior studies have adopted the pseudo-label strategies \cite{douillard2021plop, cermelli2023comformer} based on the model predictions to assign labels for seen classes while unseen classes have been ignored, thus resulting in non-discriminative features.  
\cite{joseph2021towards, truong2023fairness} improved the background modeling by using an additional prototypical representation for unknown classes.
However, these approaches consider different unknown classes as one (i.e., $N_U = 1$) resulting in non-distinguished representations of different unknown classes.
Thus, modeling function $\phi$ in Eqn. \eqref{eqn:general_clustering} without supervision of different unknown classes (i.e., $N_U > 1$) is challenging.

Although modeling $\phi$ to determine the single feature $\mathbf{f}$ belonging to the cluster $\mathbf{c}$ is challenging, 
prior studies in clustering \cite{nguyen2021clusformer, yang2020learning, yang2019learning} have suggested that determine a set of features $\{\mathbf{f}^t_i\}_{i=1}^M$ belonging to cluster $\mathbf{c}$ should be easier.
\begin{figure}[!b]
    \centering
    \vspace{-4mm}
    \includegraphics[width=1.0\linewidth]{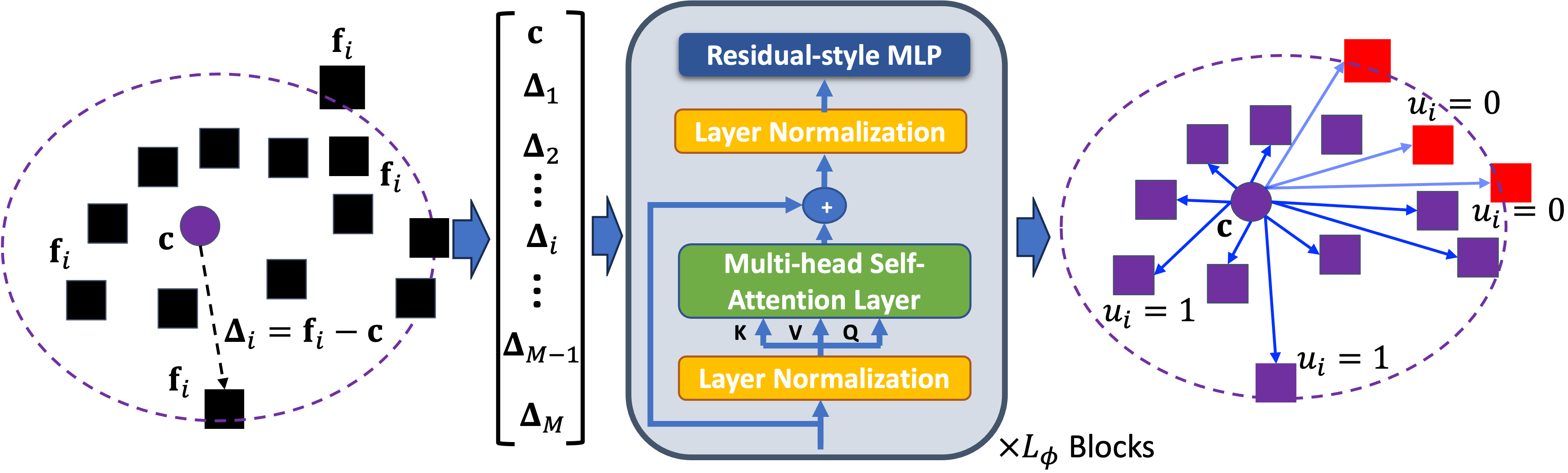}
    \vspace{-6mm}
    \caption{\textbf{The Proposed Visual Grammar Model.}}
    \label{fig:visual_grammar_model}
\end{figure}
This derives from the fact that even though the feature representations of different classes are different, \textit{the distributions of features around its cluster} (termed as \textbf{\textit{Visual Grammar}}) \textit{in the feature space should be similar among classes or clusters}.
As a result, by learning the distribution of features and their clusters, the model $\phi$ can determine whether a feature belongs to a cluster.
Then, by learning the model $\phi$ on prior known clusters and features, the knowledge of  $\phi$ can be adaptively applied to unknown clusters.
Fig. \ref{fig:visual_grammar_model} illustrates our visual grammar model of the cluster distributions.

\noindent
\textbf{Limitations of Prior Clustering Methods.}
The traditional methods in clustering, e.g., KNN or density-based clustering \cite{ester1996density}, remain limited to noisy features leading to producing the incorrect cluster assignment.
Meanwhile, the modern clustering methods, e.g., Graph Neural Networks (GNNs) \cite{yang2020learning, yang2019learning}, require a large memory to build the affinity graph for clusters. 
In addition, GNNs often learn the local structures of graphs (or clusters) and accumulate them via the aggregation layers.
Hence, the global structures of the clusters, i.e., \textit{visual grammar}, are not well modeled by GNNs \cite{nguyen2021clusformer}.
Therefore, to address these limitations, we introduced a new \textbf{\textit{Attention-based Visual Grammar}} approach to efficiently model the distribution of features and their clusters via the self-attention mechanism \cite{vaswani2017attention}.

\noindent
\textbf{Remark 1:} \textit{Given a center $\mathbf{c}$ and a set of $M$ features $\{\mathbf{f}^{\mathbf{c}}_{i}\}_{i=1}^M$ where $\mathbf{f}^{\mathbf{c}}_{i}$ denotes the feature $\mathbf{f}_{i}$ belonging to the cluster $\mathbf{c}$, and $ \forall i \in [1..M-1]: \cos(\mathbf{f}_i^{\mathbf{c}}, \mathbf{c}) \geq \cos(\mathbf{f}_{i+1}^{\mathbf{c}}, \mathbf{c})$
the \textbf{Visual Grammar} of the cluster $\mathbf{c}$ parameterized by $\Theta$ can be defined as in Eqn. \eqref{eqn:visual_grammar}.}
\begin{equation} \label{eqn:visual_grammar}
\footnotesize
\begin{split}
    \Theta^{*} &= \arg\min_{\Theta} \mathbb{E}_{\mathbf{c}, \{\mathbf{f}^{\mathbf{c}}_{i}\}_{i=1}^M} \left[-\log p(\mathbf{f}^{\mathbf{c}}_1, \mathbf{f}^{\mathbf{c}}_2, ..., \mathbf{f}^{\mathbf{c}}_M, \mathbf{c}, \Theta)\right] \\
    \Leftrightarrow \Theta^{*} &= \arg\min_{\Theta} \mathbb{E}_{\mathbf{c}, \{\mathbf{f}^{\mathbf{c}}_{i}\}_{i=1}^M} \left[-\log p(\mathbf{\Delta}^{\mathbf{c}}_1, \mathbf{\Delta}^{\mathbf{c}}_2, ..., \mathbf{\Delta}^{\mathbf{c}}_M, \mathbf{c}, \Theta)\right]
\end{split}
\end{equation}
where $\Delta^{\mathbf{c}}_i = \mathbf{f}^{\mathbf{c}}_i - \mathbf{c}$.
Eqn. \eqref{eqn:visual_grammar} defines the visual grammar of the cluster by modeling the feature distribution of $\mathbf{f}_i^{\mathbf{c}}$ and its cluster center $\mathbf{c}$.
Let $\phi: \mathbb{R}^{(M + 1) \times D} \to [0, 1]^{M}$ be a function receiving a center $\mathbf{c}$ and a set of $M$ features $\{\mathbf{f}_i\}_{i=1}^M$ $\left(\cos(\mathbf{f}_i, \mathbf{c}) \geq \cos(\mathbf{f}_{i+1}, \mathbf{c})\right)$ to determine whether $\mathbf{f}_i$ belonging to $\mathbf{c}$, i.e., 
$\mathbf{u} = \phi(\mathbf{\Delta}_1, \mathbf{\Delta}_2 , ..., \mathbf{\Delta}_{M}, \mathbf{c})$  
where $\mathbf{\Delta}_i = \mathbf{f}_i - \mathbf{c}$, $\mathbf{u} = [u_1, u_2, ..., u_M]$ and $u_i = 1$ denotes $\mathbf{f}_i$ belong to cluster $\mathbf{c}$ and vice versa. 
Hence, the visual grammar model in Eqn. \eqref{eqn:visual_grammar} can be modeled by the network $\phi$ as follows with parameter $\Theta$ as in Eqn. \eqref{eqn:visual_grammar_phi}.
\begin{equation} \label{eqn:visual_grammar_phi}
\footnotesize
\begin{split}
    \Theta^{*} &= \arg\min_{\Theta} \mathbb{E}_{\mathbf{c}, \{\mathbf{f}_{i}\}_{i=1}^M} \left[-\log p(\mathbf{u} | \mathbf{\Delta}_1, \mathbf{\Delta}_2, ..., \mathbf{\Delta}_M, \mathbf{c}, \Theta)\right]
\end{split}
\end{equation}
Eqn. \eqref{eqn:visual_grammar_phi} aims to model the distribution of features around its cluster by learning the correlation of relatively topological structures $\mathbf{\Delta}_i$ of features $\mathbf{f}_i$ around cluster $\mathbf{c}$.
Then, based on knowledge of the cluster distribution, the model $\phi$ is able to determine whether a feature $\mathbf{f}_i$ belongs to cluster $\mathbf{c}$.
Hence, it is essential that the model $\phi$ has the ability to exploit the correlation between features $\mathbf{f}_i$ and cluster $\mathbf{c}$ to learn the topological structure of visual grammar. 
Therefore, 
we adopt the self-attention mechanism \cite{vaswani2017attention, nguyen2021clusformer} to efficiently model these feature correlations.
Particularly, the model $\phi$ is formed by $L_{\phi}$ blocks of self-attention as follows,
\begin{equation}
\footnotesize
\begin{split}
    \mathbf{z}_0 &= \operatorname{LN([\mathbf{\Delta}_1, ..., \mathbf{\Delta}_M, \mathbf{c}])} + \boldsymbol{\beta}, \quad
    \mathbf{a}_l = \mathbf{z}_l + \operatorname{MHSA}(\mathbf{z}) \\ 
    \mathbf{z}_{l+1} &= \mathbf{a}_l + \operatorname{MLP}(\operatorname{LN}(\mathbf{a}_l)),  \quad\quad\quad\quad \mathbf{u} = \operatorname{Proj}(\mathbf{z}_{L_{\phi}})
\end{split}
\end{equation}
where $\boldsymbol{\beta}$ is the positional embedding, $\operatorname{LN}$ is Layer Normalization, $\operatorname{MHSA}$ is multi-head self-attention, $\operatorname{MLP}$ is the multi-layer perception, and $\operatorname{Proj}$ is the linear projection.
By using Transformers, the correlation of cluster distributions can be well modeled by the self-attention mechanism.

\noindent
\textbf{Cluster Assignment via Visual Grammar.} 
Instead of assigning the clusters based on the model prediction \cite{douillard2021plop, cermelli2023comformer, cswkd_cvpr_2022} or nearest cluster \cite{truong2023fairness, joseph2021towards} that are less effective,
the cluster assignment in our approach will be performed by the visual grammar model, i.e., the visual grammar model will consider the $M$ closest features around cluster $\mathbf{c}$ to assign the cluster for these features.
Then, the cluster assignments are used to compute our Fairness Contrastive Clustering loss. 
In addition, following common practices \cite{douillard2021plop, cermelli2023comformer, cswkd_cvpr_2022}, we improve background shift modeling by using 
the cluster assignments of features as the pseudo labels of pixels.

\noindent
\textbf{Unknown Cluster Initialization.} Prior work \cite{truong2023fairness, joseph2021towards} initialized a single unknown cluster ($N_U=1$), thus resulting in producing non-discriminative class-wise features.
However, there should be more than a single unknown cluster ($N_U>1$) to produce discriminative features for different unknown classes. 
Therefore, our approach first initializes a list of potential unknown clusters at each learning step via DB-SCAN \cite{ester1996density} on the features of unknown classes extracted by the current CSS model.
For the new known class $\mathcal{C}^{t}$, we initialize these clusters based on the mean of their feature representations. Meanwhile, the clusters of known classes learned in previous steps are maintained.
Further details can be found in the appendix.

\subsection{Continual Learning Procedure}

Fig. \ref{fig:css_framework} illustrates the training procedure of our continual learning approach. At each learning step $t$, the CSS model $F$ with $\theta_t$ is trained with the \textbf{\textit{Fairness Contrastive Clustering}} loss defined in Eqn. \eqref{eqn:loss_for_one_cluster_alpha} and the previous visual grammar model $\phi$ with $\Theta_{t-1}$. In addition, we introduce a cluster regularizer $\mathcal{R}_{C}$ to avoid the clusters of different classes collapsing into a single cluster. Therefore, the entire CSS learning objective in our approach can be formed as in Eqn. \eqref{eqn:final-equation-css}.
\begin{equation}\label{eqn:final-equation-css}
\scriptsize
\vspace{-3mm}
\begin{split}
    \arg\min_{\theta_t} \mathbb{E}_{\mathbf{x}^t, \mathbf{\hat{y}}^t} \Big[\mathcal{L}_{CE}\left(\mathbf{y}^t,  \mathbf{\hat{y}}^t\right) + \lambda_{CL}\sum_{\mathbf{c}_i}\mathcal{L}^{\alpha}_{Cont}\left(\mathbf{F}^t, \mathbf{c}_i \right)  + \lambda_{C}\mathcal{R}_{C}(\mathbf{c})\Big]
\end{split}
\raisetag{5pt}
\end{equation}
where $\mathcal{R}_{C}(\mathbf{c}) = \sum_{\mathbf{c}_i, \mathbf{c}_j}\{\max(0, 2\nabla - ||\mathbf{c}_i -\mathbf{c}_j||)\}^2$ is the regularizer to avoid the cluster collapsing,
$\lambda_C$ is the balanced weight, 
and $\nabla$ is the margin between clusters. 

\noindent
\textbf{Training Procedure of Visual Grammar Model.} At CSS learning step $t$, we adopt the visual grammar model trained on the previous learning step, i.e., $\phi$ with $\Theta_{t-1}$, to perform the cluster assignment for the contrastive clustering loss defined in Eqn. \eqref{eqn:general_clustering}.
Then, the visual grammar model  $\phi$ with $\Theta_t$ at learning step $t$ will be learned (initialized from $\Theta_{t-1}$) on the features extracted from the dataset and the set of known clusters $\mathbf{c}$ up to the current learning step. 
Following \cite{nguyen2021clusformer}, we sample a center $\mathbf{c}$ from the known clusters and its $M$ closest features to train the visual grammar model. 

\noindent
\textbf{Initial Visual Grammar Model.} At the first learning step $t=1$, since no clusters have been learned initially, the visual grammar model $\phi$ with $\Theta_{0}$ is unavailable.
However, as common practices in CSS \cite{douillard2021plop, cermelli2023comformer, ssul_neurips_2021}, the segmentation model is typically trained from a pre-trained backbone on ImageNet \cite{deng2009imagenet}.
As a result, the features extracted at the first learning step are characterized by the ImageNet features. Therefore, we adopt this philosophy to initialize our visual grammar model ($\phi$ with $\Theta_0$) by pre-training the visual grammar model on the ImageNet dataset.
Then, during CSS training, we will progressively train our visual grammar model at each learning step as aforementioned.

\section{Experiments}

\subsection{Implementations and Evaluation Protocols}

    \begin{table}[!t] %
        \centering
        \caption{Effectiveness of Fairness Contrastive Learning Loss.} \label{tab:abl_fairness_loss}
        \vspace{-3mm}
        \resizebox{1.0\linewidth}{!}{  
        \begin{tabular}{c  c | c c c c c }
        \hline
        \multicolumn{7}{c}{(a) ADE20K 100-50}                 \\
        
        \hline
        $\mathcal{L}_{Cont}$ & $\mathcal{L}^{\alpha}_{Cont}$ & 0-100 & 100-150 & all  & Major & Minor \\
        \hline
        
        \cmark &    & 44.6  & 15.2    & 34.8 & 51.5  & 26.4  \\
         & \cmark   & \textbf{44.6} & \textbf{24.5} & \textbf{37.9} & \textbf{52.1} & \textbf{30.8} \\
        
        \hline
        \multicolumn{7}{c}{(b) ADE20K 100-10}                 \\
        \hline
        
        $\mathcal{L}_{Cont}$ & $\mathcal{L}^{\alpha}_{Cont}$ & 0-100 & 100-150 & all  & Major & Minor \\
        
        \hline
        
        \cmark & &  41.9  & 16.0    & 33.2 & 49.9  & 24.9  \\
        & \cmark  &  \textbf{44.4}  & \textbf{20.4}    & \textbf{36.4} & \textbf{51.8}  & \textbf{28.7}  \\
        
        \hline
        \end{tabular}
        }
        \vspace{-6mm}
    \end{table}%
    
    \begin{table}[!b] %
        \centering
        \vspace{-6mm}
        \caption{Effectiveness of Visual Grammar.} \label{tab:abl_visual_grammar}
        \vspace{-3mm}
        \resizebox{1.0\linewidth}{!}{  
        \begin{tabular}{l| c c c c c }
        \hline
        \multicolumn{6}{c}{(a) ADE20K 100-50}                    \\
        \hline
                        & 0-100 & 101-150 & all  & Major & Minor \\
        \hline
        
        Nearest Cluster & 44.3	& 11.5	& 33.4	& 51.5	& 24.3 \\
        Fixed $\phi$       & 44.6  & 17.6    & 35.6 & 52.0  & 27.4  \\
        Adaptive $\phi$     & \textbf{44.6} & \textbf{24.5} & \textbf{37.9} & \textbf{52.1} & \textbf{30.8} \\
        
        \hline
        \multicolumn{6}{c}{(b) ADE20K 100-10}                    \\
        \hline
        
                        & 0-100 & 101-150 & all  & Major & Minor \\
        \hline

        Nearest Cluster & 40.1	& 14.3	& 31.5	& 48.7	& 22.9 \\
        Fixed $\phi$       & 43.0  & 18.5    & 34.9 & 50.6  & 27.0  \\
        Adaptive $\phi$     &  \textbf{44.4}  & \textbf{20.4}    & \textbf{36.4} & \textbf{51.8}  & \textbf{28.7}  \\
        \hline
        \end{tabular}
        }
    \end{table}

\noindent
\textbf{Implementation.}
Following common practices \cite{douillard2021plop, truong2023fairness, cermelli2020modelingthebackground}, we adopt DeepLab-V3 \cite{chen2017rethinking} with ResNet-101 \cite{He2015} and SegFormer \cite{xie2021segformer} with MiT-B3 \cite{xie2021segformer} in our experiments.
For the Visual Grammar model, we adopt the design of \cite{nguyen2021clusformer} with $L_{\phi}=12$ blocks of multi-head self-attention layers.
The feature vectors from the last layer of the decoder are used for our $\mathcal{L}^{\alpha}_{Cont}$ loss. 
The value $\alpha$ is set individually for each dataset, i.e., $\alpha = 5\times10^{-2}$ for ADE20K, $\alpha = 10^{-2}$ for VOC for Cityscapes.
The details of our hyper-parameters are provided in the appendix.

    \begin{table}[!t]
        \centering
        \caption{Effectiveness of Scaling Factor $\alpha$.} \label{tab:abl_alpha}
        \vspace{-3mm}
        \setlength{\tabcolsep}{10pt}
        \resizebox{1.0\linewidth}{!}
        {  
        \begin{tabular}{l| c c c c c } %
        \hline
        \multicolumn{6}{c}{(a) ADE20K 100-50} \\
        \hline
        $\alpha$  & 0-15 & 16-20 & all & Major & Minor \\ 
        \hline
        $\alpha = 0.1$    & 43.1 & 19.8 & 35.3 & 50.6 & 27.7 \\
        $\alpha = 0.05$   & \textbf{44.6} & \textbf{24.5} & \textbf{37.9} & \textbf{52.1} & \textbf{30.8} \\
        $\alpha = 0.01$   & 43.6 & 21.3 & 36.2 & 51.0 & 28.7 \\
        $\alpha = 0.005$  & 42.4 & 18.6 & 34.5 & 50.1 & 26.6 \\
        \hline
        \multicolumn{6}{c}{(b) Pascal VOC 15-5} \\
        \hline
        $\alpha$  & 0-15 & 16-20 & all & Major & Minor \\ 
        \hline
        $\alpha = 0.1$    & 74.8 & 51.6 & 69.3 & 76.9 & 63.5 \\
        $\alpha = 0.05$   & 76.2 & 51.3 & 70.3 & 79.0 & 63.8 \\
        $\alpha = 0.01$    & \textbf{79.4} & \textbf{54.8} & \textbf{73.5} & \textbf{81.3} & \textbf{67.7} \\
        $\alpha = 0.005 $ & 74.6 & 48.9 & 68.5 & 77.6 & 61.6 \\
        \hline
        \end{tabular}
        }
        \vspace{-6mm}
    \end{table}%
    
    \begin{table}[!b]
        \centering
        \vspace{-6mm}
        \caption{Effectiveness of Our Proposed Losses.}
        \label{tab:abl_loss_contribute}
        \vspace{-3mm}
        \resizebox{1.0\linewidth}{!}{  
        \begin{tabular}{cccc|ccccc}
        \hline
        \multicolumn{9}{c}{(a) ADE20K 100-50}                                                                                       \\
        \hline
        $\mathcal{L}_{CE}$                 & $\mathcal{L}^{\alpha}_{Cont}$ & $\phi$   & $\mathcal{R}_C$ & 0-100 & 101-150 & all  & Major & Minor \\
        \hline
        \cmark &        &        & & 0.0	& 18.9	& 6.3	& 0.0	& 9.4 \\
        \cmark & \cmark &        & & 44.0  & 7.9     & 31.9 & 51.6  & 22.1  \\
        \cmark & \cmark & \cmark & & 43.8 & 21.8 & 36.4 & 51.1 & 29.1 \\
        \cmark & \cmark & \cmark & \cmark & \textbf{44.6} & \textbf{24.5} & \textbf{37.9} & \textbf{52.1} & \textbf{30.8} \\
        \hline
        \multicolumn{9}{c}{(b) ADE20K 100-10}                                                                                       \\
        \hline
        $\mathcal{L}_{CE}$                 & $\mathcal{L}^{\alpha}_{Cont}$ & $\phi$ & $\mathcal{R}_C$ & 0-100 & 101-150 & all  & Major & Minor \\
        \hline
        \cmark &        &        & & 0.0	& 3.5	& 1.2	& 0.0	& 1.8 \\
        \cmark & \cmark &        &  & 39.0  & 13.1    & 30.4 & 47.8  & 21.6  \\
        \cmark & \cmark & \cmark & & 43.4 & 18.5 & 35.1 & 51.2 & 27.1 \\ 
        \cmark & \cmark & \cmark & \cmark & \textbf{44.4}  & \textbf{20.4}    & \textbf{36.4} & \textbf{51.8}  & \textbf{28.7}  \\
        \hline
        \end{tabular}
        }
    \end{table}

\noindent
\textbf{Evaluation Protocols.} 
We evaluate models on three standard datasets of CSS, i.e., ADE20K \cite{ade20k_challenge}, Pascal VOC \cite{Everingham15}, and Cityscapes \cite{cordts2016cityscapes}.
Following common practices \cite{truong2023fairness, cermelli2023comformer, ssul_neurips_2021}, our experiments are conducted on the overlapped CSS settings. 
In particular, on ADE20K, we use three different settings, i.e., ADE20K 100-50 (2 steps), ADE20K 100-10 (6 steps), and ADE20K 100-5 (11 steps).
On Pascal VOC, we evaluate FALCON in three benchmarks, i.e., VOC 15-5 (2 steps), VOC 15-1 (6 steps), and VOC 10-1 (11 steps).
On Cityscapes, we conduct domain incremental experiments with three settings, i.e., Cityscapes 11-5 (3 steps), Cityscapes 11-1 (11 steps), and Cityscapes 1-1 (21 steps).
Following \cite{douillard2021plop, cermelli2023comformer}, the mean Intersection over Union (mIoU) metric is adopted in our comparison, including mIoU of the last learning step on initial classes, incremental classes, and all classes. 
In addition, to illustrate the fairness improvement, we report the mIoU of major and minor classes.

\subsection{Ablation Study}

\noindent
\textbf{Effectiveness of Fairness Contrastive Clustering.}
Table \ref{tab:abl_fairness_loss} presents our results using DeepLab-V3 \cite{chen2018deeplab} with Resnet101 on ADE20K 100-50 and ADE20K 100-10 benchmarks.
We evaluate the impact of the fairness contrastive clustering loss $\mathcal{L}^{\alpha}_{Cont}$ by comparing it with the vanilla contrastive clustering loss $\mathcal{L}_{Cont}$.
As shown in our results, the overall performance has been significantly improved to $37.9\%$ and $36.4\%$ on ADE20K 100-50 and ADE20K 100-10, respectively.
In addition, the fairness has been promoted due to the mIoU improvement of minor groups.
We also study the impact of network backbones and cluster margin $\nabla$ in our appendix.

\noindent
\textbf{Effectiveness of Scaling Factor of Cluster.}
Table \ref{tab:abl_alpha} illustrates the experimental results of the impact of different scaling factor $\alpha$ on ADE20K 100-50 and Pascal VOC 15-5 benchmarks.
As shown in Table \ref{tab:abl_alpha}, when the value of scaling factor $\alpha$ gradually decreases, the performance of our proposed approach is improved accordingly since the fairness contrastive loss in Eqn \eqref{eqn:loss_for_one_cluster_alpha} tends to be more uniform across major and minor classes.
However, when the scaling factor is too small ($\alpha = 0.005$), the impact of the loss enforcement becomes weaker leading to the weaker enforcement of the fairness contrastive clustering, resulting in lower overall performance. 
In addition, we have observed that the higher the number of classes demands the higher the value of $\alpha$ since it will increase the compactness of more clusters.

\noindent
\textbf{Effectiveness of Loss Contributions.}
\noindent
Table \ref{tab:abl_loss_contribute} illustrates the contributions of proposed learning objectives.
For the model without using visual grammar, we only use a single unknown cluster ($N_U = 1$) and adopt the nearest cluster strategies to assign clusters of unknown pixels.
By using only cross-entropy loss, the mIoU performance remains low due to catastrophic forgetting and background shift problems.
Meanwhile, with our fairness clustering loss $\mathcal{L}^{\alpha}_{Cont}$, visual grammar model $\phi$, and the cluster regularizer $\mathcal{R}$, the mIoU performance has been significantly improved to $37.9\%$ and $36.4\%$ on ADE20K 100-50 and ADE20K 100-10, respectively.
Moreover, our FALCON has significantly promoted the fairness of segmentation models illustrated by the mIoU improvement of major and minor groups.

\begin{figure}[!t] 
    \centering
    \includegraphics[width=1.0\linewidth]{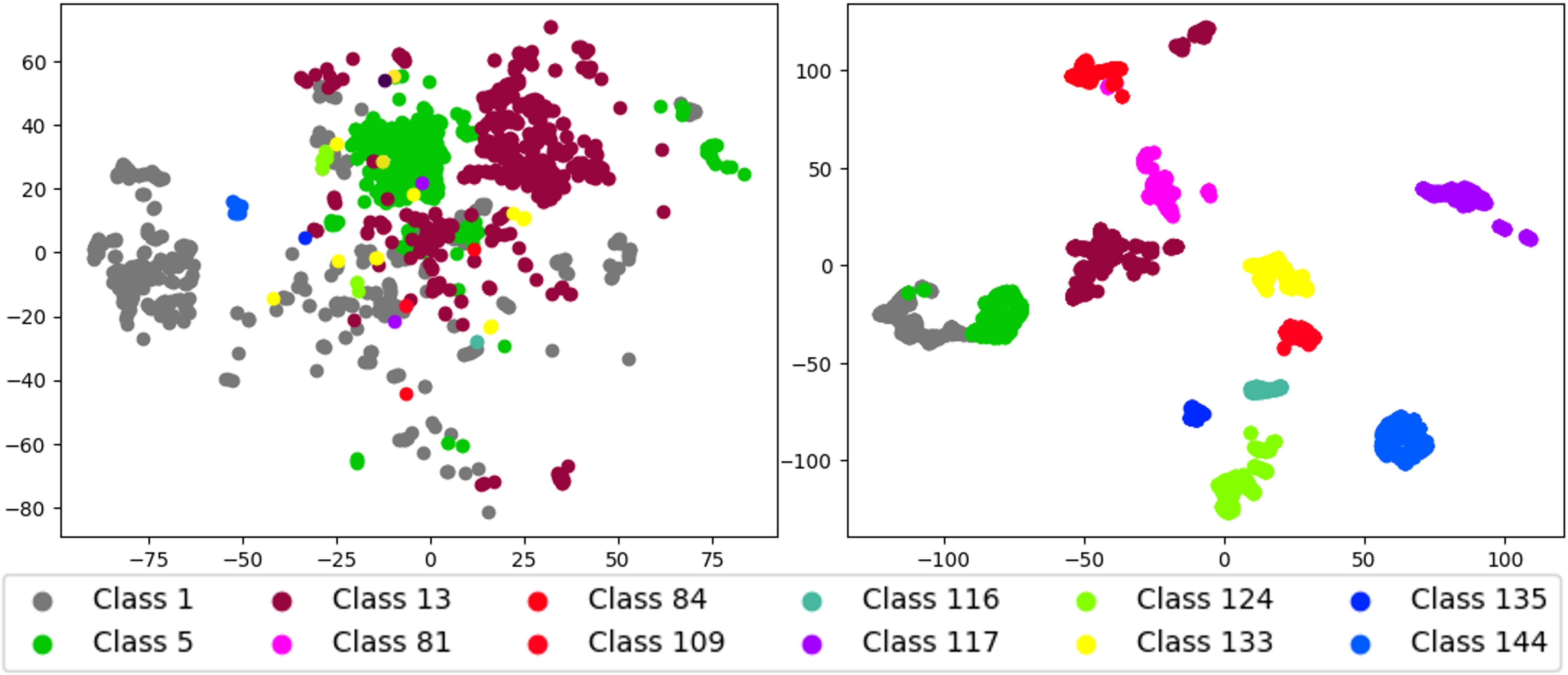}
    \vspace{-7mm}
    \caption{Cluster Distribution at Learning Step $t=1$ of ADE20K 100-50 (classes 109-144 are future classes) without (left) and with (right) Fairness Learning via Contrastive Attention.}
    \label{fig:unknown_class_distribution}
    \vspace{-6mm}
\end{figure}

\noindent
\textbf{Effectiveness of Visual Grammar.}
We evaluate FALCON under three settings, i.e., Nearest Cluster, Fixed $\phi$ pretrained ImageNet (without updating on each learning step), and Adaptive $\phi$ (with updating on each learning step).
As in Table \ref{tab:abl_visual_grammar}, the mIoU result using only the nearest cluster remains ineffective.
Meanwhile, the adaptive visual grammar model updated at each learning step further boosts the mIoU performance and promotes fairness, i.e., increased by $4.5\%$ and $4.9\%$ on ADE20K 100-50 and ADE20K 100-10 compared to the nearest cluster approach.
In addition, we study the impact of choosing the number of features $M$ in the visual grammar model in our appendix. 
Fig. \ref{fig:unknown_class_distribution} illustrates the feature distributions of unknown classes (future class).
As a result, our FALCON approach is able to model features of unknown classes into different clusters and produce better and more compact clusters compared to the one without Fairness Learning via Contrastive Attention.

\noindent
\textbf{Effectiveness af Forgetting Each Learning Step.}
Table \ref{tab:abl-each-sep} reports the results of our model evaluated at each learning step.
The performance of our model remains stable at each learning step. In particular, the maximum performance drop at learning step $4$ of classes 111-120 due forgetting is only $0.8\%$. Meanwhile, at the last learning task, the mIoU results of old classes are well maintained, with a slight drop of 0.4\% in classes 131-140.
As shown in the result, our model has effectively maintained the performance to avoid catastrophic forgetting while enabling the capability of learning classes.

\noindent
\begin{table}[!b]
\centering
\vspace{-4mm}
\caption{Effectiveness of CSS Model At Each Learning Step.}
\label{tab:abl-each-sep}
\vspace{-3mm}
\resizebox{1.0\linewidth}{!}{
\begin{tabular}{c|cccccc|c}
\hline
\multicolumn{8}{c}{ADE20K 100-10} \\
\hline
                      & 0-100  & 101-110 & 111-120 & 121-130 & 131-140 & 141-150 & all \\
\hline
Step 1                & 44.9   & $-$     & $-$     & $-$     & $-$     & $-$     & 44.9                 \\
Step 2                & 44.8   & 15.8    & $-$     & $-$     & $-$     & $-$     & 42.2                 \\
Step 3                & 44.7   & 15.7    & 22.1    & $-$     & $-$     & $-$     & 40.3                 \\
Step 4                & 44.6   & 15.6    & 21.3    & 26.3    & $-$     & $-$     & 39.2                 \\
Step 5                & 44.5   & 15.4    & 21.2    & 26.1    & 24.4    & $-$     & 37.9                 \\
Step 6                & 44.4   & 15.4    & 21.2    & 26.1    & 24.0    & 15.5    & 36.4                 \\
\hline
\multicolumn{7}{r|}{avg}                                                          & 40.1                \\
\hline
\end{tabular}
}
\end{table}

\begin{table*}[!htb]
    \begin{minipage}{.58\linewidth}
      \centering
    \caption{Comparison with Prior Methods on ADE20K Benchmarks (Note: The results of MiB \cite{cermelli2020modelingthebackground}, PLOP \cite{douillard2021plop}, and FairCL \cite{truong2023fairness} using Transformer on ADE20K 100-5 were not reported in prior studies. The upper bound results are not trained with fairness objective).}
    \label{tab:ade20k}
    \vspace{-3mm}
    \setlength{\tabcolsep}{1.5pt}
    \resizebox{1.0\linewidth}{!}{  
    \begin{tabular}{ l | l|c c c c|c c c c|c c c c }
    \hline
    \multirow{2}{*}{Network}   & \multirow{2}{*}{Method} & \multicolumn{4}{c|}{ADE20K 100-50}     & \multicolumn{4}{c|}{ADE20K 100-10}& \multicolumn{4}{c}{ADE20K 100-5} \\ \cline{3-14} 
                                &                         & 0-100  & 101-150 & all  & avg  & 0-100  & 101-150 & all  & avg  & 0-100 & 101-150 & all  & avg  \\ 
    \hline
    
    \multirow{7}{*}{DeepLab-V3} 
                                & PLOP  \cite{douillard2021plop}                  & 41.9   & 14.9    & 32.9 & 37.4 & 40.5   & 14.1    & 31.6 & 36.6 & 39.1  & 7.8     & 28.8 & 35.3 \\ %
                                & RCIL \cite{zhang2022representation}                   & 42.3   & 18.8    & 34.5 & $-$  & 39.3   & 17.6    & 32.1 & $-$  & 38.5  & 11.5    & 29.6 & $-$  \\ %
                                & REMINDER \cite{cswkd_cvpr_2022} & 41.6 & 19.2 & 34.1 & $-$ & 39.0 & 21.3 & 33.1 & $-$ & 36.1 & 16.4 & 29.5 & $-$ \\
                                & RCIL+LGKD \cite{yang2023label} & 43.3 & 25.1 & 37.2 & $-$ & 42.2 & 20.4 & 34.9 & $-$ & $-$ & $-$ & $-$ & $-$ \\
                                & FairCL \cite{truong2023fairness}                 & 43.4   & {24.0}    & 37.0 & 40.5 & 41.7   & \textbf{20.4}    & 34.7 & 39.0 & $-$   & $-$     & $-$  & $-$  \\ %
                                & \textbf{FALCON} & \textbf{44.6}	& \textbf{24.5}	& \textbf{37.9} & \textbf{41.3}  & \textbf{44.4}   & \textbf{20.4}    & \textbf{36.4} &  \textbf{40.1}    & \textbf{38.0}  & \textbf{16.1}    & \textbf{30.7} & \textbf{37.6}  \\ 
                                \cdashline{2-14} 
                                & Upper Bound & 44.3 & 28.2 & 38.9 & $-$ & 44.3 & 28.2 & 38.9 & $-$ & 44.3 & 28.2 & 38.9 & $-$ \\
                                \hline

    \multirow{5}{*}{Mask2Former} 
        & MiB \cite{cermelli2020modelingthebackground}                    & 37.0   & 24.1    & 32.6 & 38.3 & 23.5   & 10.6    & 26.6 & 29.6 & 21.0  & 6.1     & 16.1 & 27.7 \\ %
         & PLOP \cite{douillard2021plop}                   & 44.2   & 26.2    & 38.2 & 41.1 & 34.8   & 15.9    & 28.5 & 35.2 & 33.6  & 14.1    & 27.1 & 33.6 \\ %
         & CoMFormer \cite{cermelli2023comformer}               & 44.7   & 26.2    & 38.4 & 41.2 & 40.6   & 15.6    & 32.3 & 37.4 & 39.5  & 13.6    & 30.9 & 36.5 \\ %
         & ECLIPSE \cite{kim2024eclipse} & 45.0 & 21.7 & 37.1 & $-$ & 43.4 & 17.4 & 34.6 & $-$ & \textbf{43.3} & 16.3 & \textbf{34.2} & $-$ \\
         & CoMasTRe \cite{gong2024continual} & 45.7 & 26.0 & 39.2 & 41.6 & 42.3 & 18.4 & 34.4 & 38.4 & 40.8 & 15.8 & 32.6 & \textbf{38.6} \\
         \hline 
         
         & MiB \cite{cermelli2020modelingthebackground} &  43.4 & \textbf{30.6} & 39.2 & 38.7 & 39.1 & 20.4 & 34.2 & 39.5 & $-$ & $-$ & $-$ & $-$ \\
         & PLOP \cite{douillard2021plop} & 43.8 & 26.2 & 38.0 & 38.1 & 43.3 & 24.1 & 36.2 & 40.3 & $-$ & $-$ & $-$ & $-$ \\
        Transformer & FairCL \cite{truong2023fairness}                 & 43.6 & 25.5 & 37.6 & 40.7 & 42.2 & 21.9 & 35.5 & 39.4 & $-$   & $-$ & $-$  & $-$ \\
         & \textbf{FALCON}                    & \textbf{47.5}   & \textbf{30.6}    & \textbf{41.9} & \textbf{43.5}  & \textbf{47.3}   & \textbf{26.2}    & \textbf{40.3} & \textbf{42.8}  & {40.8}  & \textbf{18.9}    & {33.5} & {38.1}  \\
    \cdashline{2-14}
    & Upper Bound & 48.7 & 39.0 & 45.5 & $-$ & 48.7 & 39.0 & 45.5 & $-$ & 48.7 & 39.0 & 45.5 & $-$ \\
    
    \hline
    \end{tabular}
    }
    \end{minipage}%
    \hfill
    \begin{minipage}{.4\linewidth}
    \centering
    \includegraphics[width=1.0\linewidth]{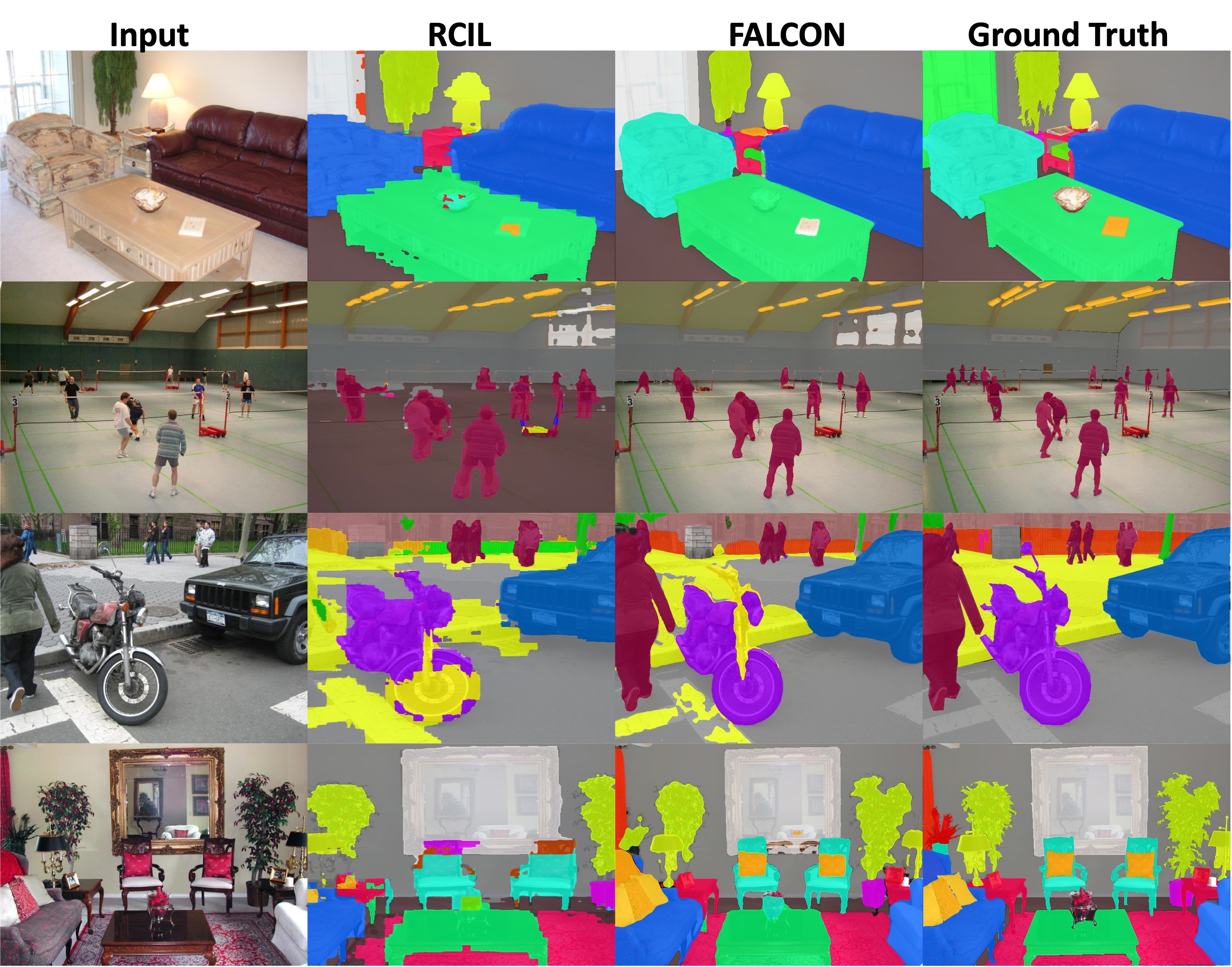}
    \vspace{-7mm}
    \captionof{figure}{The Comparison of Qualitative Results on ADE20K Between Our Approach and RCIL \cite{zhang2022representation}.}
    \label{fig:visualization}
    \end{minipage} 
    \vspace{-6mm}
\end{table*}

\begin{table*}[!b]
\vspace{-4mm}
	\begin{minipage}[t]{0.60\textwidth}
	   \centering
\caption{Comparisons with Prior Methods on Pascal VOC.}
\label{tab:voc}
\vspace{-3mm}
\setlength{\tabcolsep}{4pt}
\resizebox{1.0\textwidth}{!}{  
\begin{tabular}{l | l|ccc|ccc|ccc}
\hline
\multirow{2}{*}{}  & \multirow{2}{*}{Method} & \multicolumn{3}{c|}{Pascal VOC 15-5} & \multicolumn{3}{c|}{Pascal VOC 15-1} & \multicolumn{3}{|c}{Pascal VOC 10-1 } \\
\cline{3-11}
                       & & 0-15       & 16-20     & all       & 0-15       & 16-20     & all       & 0-10       & 11-20      & all       \\
\hline
\multirow{7}{*}{\rot{DeepLab-V3}} & MiB \cite{cermelli2020modelingthebackground}             & 76.37      & 49.97     & 70.08     & 38.00      & 13.50     & 32.20     & 20.00      & 20.10      & 20.10     \\
 & PLOP \cite{douillard2021plop}           & 75.73      & 51.71     & 70.09     & 65.10      & 21.10     & 54.60     & 44.00      & 15.50      & 30.50     \\
& RCIL \cite{zhang2022representation}           & $-$        & $-$       & $-$       & 70.60      & 23.70     & 59.40     & 55.40      & 15.10      & 34.30     \\
& FairCL  \cite{truong2023fairness}     & $-$        & $-$       & $-$       & 72.00      & 22.70     & 60.30     & 42.30      & 25.60      & 34.40     \\
& SSUL  \cite{ssul_neurips_2021}                  & 77.82      & 50.10     & 71.22     & 77.31      & 36.59     & 67.61     & 71.31      & 45.98      & 59.25     \\
& \textbf{FALCON}        & 
\textbf{79.35}      & \textbf{54.77}     & \textbf{73.50}     & \textbf{78.34}      & \textbf{42.57}     & \textbf{69.83}     &   \textbf{73.94}	& \textbf{49.73}	& \textbf{62.41}\\
\cdashline{2-11}
& Upper Bound                    & 79.77      & 72.35     & 77.43     & 79.77      & 72.35     & 77.43     & 78.41      & 76.35      & 77.43    \\
\hline
\multirow{5}{*}{\rot{Transformer}} & PLOP \cite{douillard2021plop} & 72.51 & 48.37 & 66.76 & 64.59 & 37.23 & 58.08 & 48.53 & 33.71 & 41.47 \\
& SSUL \cite{ssul_neurips_2021} & 79.91 & {56.83} & {74.41} & \textbf{79.91} & 40.56 & 70.54 & 74.06 & 51.85 & 63.48 \\
 & FairCL   \cite{truong2023fairness}    & $-$        & $-$       & $-$       & 73.50      & 22.80     & 61.50     & 57.10      & 14.20      & 36.60     \\
 & \textbf{FALCON} & \textbf{81.20} &	\textbf{58.04} &	\textbf{75.69} & {78.71}      & \textbf{47.54}     & \textbf{71.28}     & \textbf{74.92}	& \textbf{52.54}	& \textbf{64.26}     \\
\cdashline{2-11}
& Upper Bound & 80.84 & 74.97 & 79.44 & 80.84 & 74.97 & 79.44 & 80.84 & 74.97 & 79.44 \\ 
\hline
\end{tabular}
}
	   
	\end{minipage} 
	\hspace{0.5mm}
	\begin{minipage}[t]{0.4\textwidth}

 \centering
\caption{Comparisons on Cityscapes.}
\label{tab:cityscape}
\vspace{-3mm}
\setlength{\tabcolsep}{5pt}
\resizebox{0.92\textwidth}{!}{  
\begin{tabular}{l | l |c c c}
\hline
\multirow{1}{*}{} & \multirow{1}{*}{Method} & 11-5    & 11-1     & 1-1      \\ 
\cline{1-5} 
\hline
\multirow{8}{*}{\rot{DeepLab-V3}} & LWF-MC \cite{rebuffi2017icarl}         & 58.90   & 56.92    & 31.24    \\ %
 & ILT \cite{michieli2019ilt}            & 59.14   & 57.75    & 30.11    \\ %
& MİB \cite{cermelli2020modelingthebackground}             & 61.51   & 60.02    & 42.15    \\ %
 & PLOP \cite{douillard2021plop}           & 63.51   & 62.05    & 45.24    \\ %
 & RCIL \cite{zhang2022representation}           & 64.30   & 63.00    & 48.90    \\ %
& FairCL \cite{truong2023fairness}     & 66.96   & 66.61    & 49.22    \\ %
& \textbf{FALCON}         & \textbf{70.74}   & \textbf{69.75}    & \textbf{55.24}    \\ %
\cdashline{2-5}
& Upper Bound & 79.30 & 79.30 & 79.30 \\
\hline
 \multirow{3}{*}{\rot{Trans.}} & FairCL \cite{truong2023fairness}      & 67.85   & 67.09    & 55.68    \\ %
 & \textbf{FALCON}       & \textbf{71.33}   & \textbf{70.14}    & \textbf{58.79}    \\ 
\cdashline{2-5}
& Upper Bound  & 83.80 & 83.80 & 83.80 \\
\hline
\end{tabular}
}
   	
	\end{minipage}
\end{table*}

\vspace{-4mm}
\subsection{Comparison with Prior SoTA Methods}

\noindent
\textbf{ADE20K.}
Table \ref{tab:ade20k} presents our experimental results 
compared to prior CSS methods. 
By using DeepLab-V3, our approach has achieved SoTA performance, i.e., the mIoU results of $37.9\%$ and $+36.4\%$ on ADE20K 100-50 and ADE20K 100-10 benchmarks, higher than prior FairCL \cite{truong2023fairness}. 
Meanwhile, FALCON using Transformer has outperformed the prior CoMFormer \cite{cermelli2023comformer} model by $+3.5\%$, $+8.0\%$, and $+2.6\%$ on ADE20K 100-50, ADE20K 100-10, and ADE20K 100-5.
In addition, our mIoU results on the initial classes remain competitive with the upper-bounded results because FALCON can well handle the fairness problem compared to fully supervised learning.
Although ECLIPSE \cite{kim2024eclipse} exhibits slightly better on ADE20K 100-5 by only $0.7\%$ (all) due to the better segmentation framework (i.e., Mask2Former), FALCON significantly outperforms ECLIPSE on ADE20K 100-50 and ADE20K 100-10.
We also report our results of ADE20K 50-50 in the appendix. 
As in Fig. \ref{fig:visualization}, FALCON produces better segmentation maps than prior methods.

\noindent
\textbf{Pascal VOC.}
Table \ref{tab:voc} presents our results on Pascal VOC benchmarks.
FALCON has consistently achieved the SoTA performance on three benchmarks. In particular, compared to the prior FairCL \cite{truong2023fairness} approach, our methods using DeepLab-V3 have improved the mIoU performance up to $73.50\%$, $69.83\%$, and $62.41\%$ on  Pascal VOC 15-5, Pascal VOC 15-1, and Pascal VOC 10-1, respectively. Additionally, by using the better backbone, the performance is further improved, and the gap with the upper-bound result is reduced.

\noindent
\textbf{Cityscapes.}
Table \ref{tab:cityscape} reports the performance of our approach using DeepLab-V3 compared to prior methods on three different settings, i.e., Cityscapes 11-5, Cityscapes 11-1, and Cityscapes 1-1.
As in Table \ref{tab:cityscape}, the performance of our methods has consistently outperformed prior FairCL \cite{truong2023fairness} approach by $+3.78\%$, $+3.14\%$, and $+6.02\%$ on three benchmarks. Similar to our experiments on ADE20K and VOC, the better network brings higher results.

\vspace{-3mm}
\section{Conclusions and Limitations}
\vspace{-2mm}

\textbf{Conclusions.} This paper has presented a novel Fairness Learning via Contrastive Attention approach to CSS. In particular, the fairness contrastive clustering loss has been introduced to model both catastrophic forgetting and fairness problems. Then, the visual grammar model was presented to model the unknown classes.
The experimental results on different benchmarks have shown the SoTA performance and fairness promotion of our proposed FALCON approach.

\noindent
\textbf{Limitations.} Our study chose a set of learning hyper-parameters to support our theoretical analysis.
However, it potentially consists of several limitations related to choosing learning parameters and cluster initialization. 
Further discussion of limitations is presented in the supplementary.

{
\noindent
\textbf{Acknowledgment} 
This material is based upon work supported by the National Science Foundation under Award No. OIA-1946391.
We also acknowledge the Arkansas High-Performance Computing Center for providing GPUs.
}

{
\small
\bibliographystyle{ieeenat_fullname}
\bibliography{references}

\begin{thebibliography}{78}
\providecommand{\natexlab}[1]{#1}
\providecommand{\url}[1]{\texttt{#1}}
\expandafter\ifx\csname urlstyle\endcsname\relax
  \providecommand{\doi}[1]{doi: #1}\else
  \providecommand{\doi}{doi: \begingroup \urlstyle{rm}\Url}\fi

\bibitem[Araslanov et~al.(2021)Araslanov, , and Roth]{Araslanov:2021:DASAC}
Nikita Araslanov, , and Stefan Roth.
\newblock Self-supervised augmentation consistency for adapting semantic
  segmentation.
\newblock In \emph{Proceedings of the IEEE Conference on Computer Vision and
  Pattern Recognition (CVPR)}, 2021.

\bibitem[Bottou(2010)]{bottou2010large}
L{\'e}on Bottou.
\newblock Large-scale machine learning with stochastic gradient descent.
\newblock In \emph{COMPSTAT}, 2010.

\bibitem[Cermelli et~al.(2020)Cermelli, Mancini, Rota~Bulò, Ricci, and
  Caputo]{cermelli2020modelingthebackground}
Fabio Cermelli, Massimiliano Mancini, Samuel Rota~Bulò, Elisa Ricci, and
  Barbara Caputo.
\newblock Modeling the background for incremental learning in semantic
  segmentation.
\newblock In \emph{Proc. Conf. Comp. Vision Pattern Rec.}, 2020.

\bibitem[Cermelli et~al.(2023)Cermelli, Cord, and
  Douillard]{cermelli2023comformer}
Fabio Cermelli, Matthieu Cord, and Arthur Douillard.
\newblock Comformer: Continual learning in semantic and panoptic segmentation.
\newblock \emph{IEEE/CVF Computer Vision and Pattern Recognition Conference},
  2023.

\bibitem[Cha et~al.(2021)Cha, kim, Yoo, and Moon]{ssul_neurips_2021}
Sungmin Cha, beomyoung kim, YoungJoon Yoo, and Taesup Moon.
\newblock Ssul: Semantic segmentation with unknown label for exemplar-based
  class-incremental learning.
\newblock In \emph{Advances in Neural Information Processing Systems}, pages
  10919--10930. Curran Associates, Inc., 2021.

\bibitem[Chen et~al.(2024)Chen, Cong, Luo, Ip, and Kwong]{chen2024strike}
Jinpeng Chen, Runmin Cong, Yuxuan Luo, Horace Ho~Shing Ip, and Sam Kwong.
\newblock Strike a balance in continual panoptic segmentation.
\newblock \emph{arXiv preprint arXiv:2407.16354}, 2024.

\bibitem[Chen et~al.(2017)Chen, Papandreou, Schroff, and
  Adam]{chen2017rethinking}
Liang-Chieh Chen, George Papandreou, Florian Schroff, and Hartwig Adam.
\newblock Rethinking atrous convolution for semantic image segmentation.
\newblock \emph{arXiv preprint arXiv:1706.05587}, 2017.

\bibitem[Chen et~al.(2018)Chen, Papandreou, Kokkinos, Murphy, and
  Yuille]{chen2018deeplab}
Liang-Chieh Chen, George Papandreou, Iasonas Kokkinos, Kevin Murphy, and Alan~L
  Yuille.
\newblock Deeplab: Semantic image segmentation with deep convolutional nets,
  atrous convolution, and fully connected {CRF}s.
\newblock \emph{TPAMI}, 2018.

\bibitem[Chen et~al.(2022)Chen, Fu, Narayan, Zhang, Song, Fatahalian, and
  R{\'e}]{chen2022perfectly}
Mayee Chen, Daniel~Y Fu, Avanika Narayan, Michael Zhang, Zhao Song, Kayvon
  Fatahalian, and Christopher R{\'e}.
\newblock Perfectly balanced: Improving transfer and robustness of supervised
  contrastive learning.
\newblock In \emph{International Conference on Machine Learning}, pages
  3090--3122. PMLR, 2022.

\bibitem[Chen et~al.(2020{\natexlab{a}})Chen, Kornblith, Norouzi, and
  Hinton]{chen2020simple}
Ting Chen, Simon Kornblith, Mohammad Norouzi, and Geoffrey Hinton.
\newblock A simple framework for contrastive learning of visual
  representations.
\newblock In \emph{International conference on machine learning}, pages
  1597--1607. PMLR, 2020{\natexlab{a}}.

\bibitem[Chen et~al.(2020{\natexlab{b}})Chen, Fan, Girshick, and
  He]{chen2020improved}
Xinlei Chen, Haoqi Fan, Ross Girshick, and Kaiming He.
\newblock Improved baselines with momentum contrastive learning.
\newblock \emph{arXiv preprint arXiv:2003.04297}, 2020{\natexlab{b}}.

\bibitem[Chen* et~al.(2021)Chen*, Xie*, and He]{chen2021mocov3}
Xinlei Chen*, Saining Xie*, and Kaiming He.
\newblock An empirical study of training self-supervised vision transformers.
\newblock In \emph{ICCV}, 2021.

\bibitem[Cheng et~al.(2021)Cheng, Schwing, and Kirillov]{cheng2021maskformer}
Bowen Cheng, Alexander~G. Schwing, and Alexander Kirillov.
\newblock Per-pixel classification is not all you need for semantic
  segmentation.
\newblock In \emph{NeurIPS}, 2021.

\bibitem[Cheng et~al.(2022)Cheng, Misra, Schwing, Kirillov, and
  Girdhar]{cheng2022masked}
Bowen Cheng, Ishan Misra, Alexander~G Schwing, Alexander Kirillov, and Rohit
  Girdhar.
\newblock Masked-attention mask transformer for universal image segmentation.
\newblock In \emph{Proceedings of the IEEE/CVF conference on computer vision
  and pattern recognition}, pages 1290--1299, 2022.

\bibitem[Chu et~al.(2021)Chu, Kim, and Han]{chu2021learning}
Sanghyeok Chu, Dongwan Kim, and Bohyung Han.
\newblock Learning debiased and disentangled representations for semantic
  segmentation.
\newblock \emph{Advances in Neural Information Processing Systems},
  34:\penalty0 8355--8366, 2021.

\bibitem[Cong et~al.(2024)Cong, Cong, Liu, and Sun]{cong2025cs}
Wei Cong, Yang Cong, Yuyang Liu, and Gan Sun.
\newblock Cs2k: Class-specific and class-shared knowledge guidance for
  incremental semantic segmentation.
\newblock In \emph{European Conference on Computer Vision}, pages 244--261.
  Springer, 2024.

\bibitem[Cordts et~al.(2016)Cordts, Omran, Ramos, Rehfeld, Enzweiler, Benenson,
  Franke, Roth, and Schiele]{cordts2016cityscapes}
Marius Cordts, Mohamed Omran, Sebastian Ramos, Timo Rehfeld, Markus Enzweiler,
  Rodrigo Benenson, Uwe Franke, Stefan Roth, and Bernt Schiele.
\newblock The {C}ityscapes dataset for semantic urban scene understanding.
\newblock In \emph{CVPR}, 2016.

\bibitem[Cui et~al.(2021)Cui, Zhong, Liu, Yu, and Jia]{cui2021parametric}
Jiequan Cui, Zhisheng Zhong, Shu Liu, Bei Yu, and Jiaya Jia.
\newblock Parametric contrastive learning.
\newblock In \emph{Proceedings of the IEEE/CVF international conference on
  computer vision}, pages 715--724, 2021.

\bibitem[Deng et~al.(2009)Deng, Dong, Socher, Li, Li, and
  Fei-Fei]{deng2009imagenet}
Jia Deng, Wei Dong, Richard Socher, Li-Jia Li, Kai Li, and Li Fei-Fei.
\newblock Imagenet: A large-scale hierarchical image database.
\newblock In \emph{CVPR}, 2009.

\bibitem[Douillard et~al.(2020)Douillard, Cord, Ollion, Robert, and
  Valle]{douillard2020podnet}
Arthur Douillard, Matthieu Cord, Charles Ollion, Thomas Robert, and Eduardo
  Valle.
\newblock Podnet: Pooled outputs distillation for small-tasks incremental
  learning.
\newblock In \emph{Proc. Europ. Conf. Computer Vision}, 2020.

\bibitem[Douillard et~al.(2021)Douillard, Chen, Dapogny, and
  Cord]{douillard2021plop}
Arthur Douillard, Yifu Chen, Arnaud Dapogny, and Matthieu Cord.
\newblock Plop: Learning without forgetting for continual semantic
  segmentation.
\newblock In \emph{Proceedings of the IEEE/CVF Conference on Computer Vision
  and Pattern Recognition}, pages 4040--4050, 2021.

\bibitem[Ermis et~al.(2022)Ermis, Zappella, Wistuba, Rawal, and
  Archambeau]{Ermis_2022_CVPR}
Beyza Ermis, Giovanni Zappella, Martin Wistuba, Aditya Rawal, and C\'edric
  Archambeau.
\newblock Continual learning with transformers for image classification.
\newblock In \emph{Proceedings of the IEEE/CVF Conference on Computer Vision
  and Pattern Recognition (CVPR) Workshops}, pages 3774--3781, 2022.

\bibitem[Ester et~al.(1996)Ester, Kriegel, Sander, Xu,
  et~al.]{ester1996density}
Martin Ester, Hans-Peter Kriegel, J{\"o}rg Sander, Xiaowei Xu, et~al.
\newblock A density-based algorithm for discovering clusters in large spatial
  databases with noise.
\newblock In \emph{kdd}, pages 226--231, 1996.

\bibitem[Everingham et~al.(2015)Everingham, Eslami, Van~Gool, Williams, Winn,
  and Zisserman]{Everingham15}
M. Everingham, S.~M.~A. Eslami, L. Van~Gool, C.~K.~I. Williams, J. Winn, and A.
  Zisserman.
\newblock The pascal visual object classes challenge: A retrospective.
\newblock \emph{IJCV}, 2015.

\bibitem[French(1999)]{french1999catastrophicforgetting}
Robert French.
\newblock Catastrophic forgetting in connectionist networks.
\newblock \emph{Trends in Cognitive Sciences}, 1999.

\bibitem[Gong et~al.(2024)Gong, Yu, Wang, and Xiao]{gong2024continual}
Yizheng Gong, Siyue Yu, Xiaoyang Wang, and Jimin Xiao.
\newblock Continual segmentation with disentangled objectness learning and
  class recognition.
\newblock In \emph{Proceedings of the IEEE/CVF Conference on Computer Vision
  and Pattern Recognition}, pages 3848--3857, 2024.

\bibitem[He et~al.(2016)He, Zhang, Ren, and Sun]{He2015}
Kaiming He, Xiangyu Zhang, Shaoqing Ren, and Jian Sun.
\newblock Deep residual learning for image recognition.
\newblock In \emph{CVPR}, 2016.

\bibitem[He et~al.(2020)He, Fan, Wu, Xie, and Girshick]{he2020momentum}
Kaiming He, Haoqi Fan, Yuxin Wu, Saining Xie, and Ross Girshick.
\newblock Momentum contrast for unsupervised visual representation learning.
\newblock In \emph{Proceedings of the IEEE/CVF conference on computer vision
  and pattern recognition}, pages 9729--9738, 2020.

\bibitem[Hoyer et~al.(2022)Hoyer, Dai, and Van~Gool]{daformer}
Lukas Hoyer, Dengxin Dai, and Luc Van~Gool.
\newblock {DAFormer}: Improving network architectures and training strategies
  for domain-adaptive semantic segmentation.
\newblock In \emph{CVPR}, 2022.

\bibitem[Jalata et~al.(2022)Jalata, Chappa, Truong, Helton, Rainwater, and
  Luu]{jalata2022eqadap}
Ibsa Jalata, Naga Venkata Sai~Raviteja Chappa, Thanh-Dat Truong, Pierce Helton,
  Chase Rainwater, and Khoa Luu.
\newblock Eqadap: Equipollent domain adaptation approach to image deblurring.
\newblock \emph{IEEE Access}, 10:\penalty0 93203--93211, 2022.

\bibitem[Joseph et~al.(2021)Joseph, Khan, Khan, and
  Balasubramanian]{joseph2021towards}
KJ Joseph, Salman Khan, Fahad~Shahbaz Khan, and Vineeth~N Balasubramanian.
\newblock Towards open world object detection.
\newblock In \emph{Proceedings of the IEEE/CVF conference on computer vision
  and pattern recognition}, pages 5830--5840, 2021.

\bibitem[Kim et~al.(2024)Kim, Yu, and Hwang]{kim2024eclipse}
Beomyoung Kim, Joonsang Yu, and Sung~Ju Hwang.
\newblock Eclipse: Efficient continual learning in panoptic segmentation with
  visual prompt tuning.
\newblock In \emph{Proceedings of the IEEE/CVF Conference on Computer Vision
  and Pattern Recognition}, pages 3346--3356, 2024.

\bibitem[Kirkpatrick et~al.(2017)Kirkpatrick, Pascanu, Rabinowitz, Veness,
  Desjardins, Rusu, Milan, Quan, Ramalho, Grabska-Barwinska, Hassabis, Clopath,
  Kumaran, and Hadsell]{kirkpatrick2017ewc}
James Kirkpatrick, Razvan Pascanu, Neil Rabinowitz, Joel Veness, Guillaume
  Desjardins, Andrei~A. Rusu, Kieran Milan, John Quan, Tiago Ramalho, Agnieszka
  Grabska-Barwinska, Demis Hassabis, Claudia Clopath, Dharshan Kumaran, and
  Raia Hadsell.
\newblock Overcoming catastrophic forgetting in neural networks.
\newblock \emph{PNAS}, 2017.

\bibitem[Lei et~al.(2021)Lei, Li, Zhou, Gan, Berg, Bansal, and
  Liu]{lei2021less}
Jie Lei, Linjie Li, Luowei Zhou, Zhe Gan, Tamara~L Berg, Mohit Bansal, and
  Jingjing Liu.
\newblock Less is more: Clipbert for video-and-language learning via sparse
  sampling.
\newblock In \emph{Proceedings of the IEEE/CVF conference on computer vision
  and pattern recognition}, pages 7331--7341, 2021.

\bibitem[Li et~al.(2021)Li, Hu, Liu, Peng, Zhou, and Peng]{li2021contrastive}
Yunfan Li, Peng Hu, Zitao Liu, Dezhong Peng, Joey~Tianyi Zhou, and Xi Peng.
\newblock Contrastive clustering.
\newblock In \emph{Proceedings of the AAAI conference on artificial
  intelligence}, pages 8547--8555, 2021.

\bibitem[Liu et~al.(2024)Liu, Rizzoli, Zanuttigh, Li, and Niu]{liu2024learning}
Chang Liu, Giulia Rizzoli, Pietro Zanuttigh, Fu Li, and Yi Niu.
\newblock Learning from the web: Language drives weakly-supervised incremental
  learning for semantic segmentation.
\newblock \emph{arXiv preprint arXiv:2407.13363}, 2024.

\bibitem[Liu et~al.(2019)Liu, Miao, Zhan, Wang, Gong, and
  Yu]{liu2019largescale}
Ziwei Liu, Zhongqi Miao, Xiaohang Zhan, Jiayun Wang, Boqing Gong, and Stella~X.
  Yu.
\newblock Large-scale long-tailed recognition in an open world, 2019.

\bibitem[Lopez-Paz and Ranzato(2017)]{lopezpaz2017gem}
David Lopez-Paz and Marc'Aurelio Ranzato.
\newblock Gradient episodic memory for continual learning.
\newblock In \emph{NeurIPS}, 2017.

\bibitem[Michieli and Zanuttigh(2019{\natexlab{a}})]{michieli2019ilt}
Umberto Michieli and Pietro Zanuttigh.
\newblock Incremental learning techniques for semantic segmentation.
\newblock In \emph{ICCVWS}, 2019{\natexlab{a}}.

\bibitem[Michieli and Zanuttigh(2019{\natexlab{b}})]{michieli2019incremental}
Umberto Michieli and Pietro Zanuttigh.
\newblock Incremental learning techniques for semantic segmentation.
\newblock In \emph{Proceedings of the IEEE/CVF international conference on
  computer vision workshops}, pages 0--0, 2019{\natexlab{b}}.

\bibitem[Nguyen et~al.(2022)Nguyen, Truong, Huang, Liang, Le, and
  Luu]{nguyen2022self}
Pha Nguyen, Thanh-Dat Truong, Miaoqing Huang, Yi Liang, Ngan Le, and Khoa Luu.
\newblock Self-supervised domain adaptation in crowd counting.
\newblock In \emph{2022 IEEE international conference on image processing
  (ICIP)}, pages 2786--2790. IEEE, 2022.

\bibitem[Nguyen et~al.(2021)Nguyen, Bui, Duong, Bui, and
  Luu]{nguyen2021clusformer}
Xuan-Bac Nguyen, Duc~Toan Bui, Chi~Nhan Duong, Tien~D Bui, and Khoa Luu.
\newblock Clusformer: A transformer based clustering approach to unsupervised
  large-scale face and visual landmark recognition.
\newblock In \emph{Proceedings of the IEEE/CVF Conference on Computer Vision
  and Pattern Recognition}, pages 10847--10856, 2021.

\bibitem[Oord et~al.(2018)Oord, Li, and Vinyals]{oord2018representation}
Aaron van~den Oord, Yazhe Li, and Oriol Vinyals.
\newblock Representation learning with contrastive predictive coding.
\newblock \emph{arXiv preprint arXiv:1807.03748}, 2018.

\bibitem[Ozdemir and Goksel(2019)]{ozdemir2019extending}
Firat Ozdemir and Orcun Goksel.
\newblock Extending pretrained segmentation networks with additional anatomical
  structures.
\newblock \emph{International journal of computer assisted radiology and
  surgery}, 14:\penalty0 1187--1195, 2019.

\bibitem[Ozdemir et~al.(2018)Ozdemir, Fuernstahl, and Goksel]{ozdemir2018learn}
Firat Ozdemir, Philipp Fuernstahl, and Orcun Goksel.
\newblock Learn the new, keep the old: Extending pretrained models with new
  anatomy and images.
\newblock In \emph{Medical Image Computing and Computer Assisted
  Intervention--MICCAI 2018: 21st International Conference, Granada, Spain,
  September 16-20, 2018, Proceedings, Part IV 11}, pages 361--369. Springer,
  2018.

\bibitem[Phan et~al.(2022)Phan, Ta, Phung, Tran-Thanh, and
  Bouzerdoum]{cswkd_cvpr_2022}
Minh~Hieu Phan, The-Anh Ta, Son~Lam Phung, Long Tran-Thanh, and Abdesselam
  Bouzerdoum.
\newblock Class similarity weighted knowledge distillation for continual
  semantic segmentation.
\newblock In \emph{Proceedings of the IEEE/CVF Conference on Computer Vision
  and Pattern Recognition (CVPR)}, pages 16866--16875, 2022.

\bibitem[Qiu et~al.(2023)Qiu, Shen, Sun, Zheng, Chang, Zheng, and
  Wang]{sats_prj_2023}
Yiqiao Qiu, Yixing Shen, Zhuohao Sun, Yanchong Zheng, Xiaobin Chang, Weishi
  Zheng, and Ruixuan Wang.
\newblock Sats: Self-attention transfer for continual semantic segmentation.
\newblock \emph{Pattern Recognition}, 138:\penalty0 109383, 2023.

\bibitem[Radford et~al.(2021)Radford, Kim, Hallacy, Ramesh, Goh, Agarwal,
  Sastry, Askell, Mishkin, Clark, et~al.]{radford2021learning}
Alec Radford, Jong~Wook Kim, Chris Hallacy, Aditya Ramesh, Gabriel Goh,
  Sandhini Agarwal, Girish Sastry, Amanda Askell, Pamela Mishkin, Jack Clark,
  et~al.
\newblock Learning transferable visual models from natural language
  supervision.
\newblock In \emph{International conference on machine learning}, pages
  8748--8763. PMLR, 2021.

\bibitem[Rebuffi et~al.(2017)Rebuffi, Kolesnikov, Sperl, and
  Lampert]{rebuffi2017icarl}
Sylvestre-Alvise Rebuffi, Alexander Kolesnikov, Georg Sperl, and Christoph~H.
  Lampert.
\newblock icarl: Incremental classifier and representation learning.
\newblock In \emph{Proc. Conf. Comp. Vision Pattern Rec.}, 2017.

\bibitem[Ren et~al.(2020)Ren, Yu, Sheng, Ma, Zhao, Yi, and Li]{ren2020balanced}
Jiawei Ren, Cunjun Yu, Shunan Sheng, Xiao Ma, Haiyu Zhao, Shuai Yi, and
  Hongsheng Li.
\newblock Balanced meta-softmax for long-tailed visual recognition, 2020.

\bibitem[Robins(1995)]{robins1995catastrophicforgetting}
Anthony Robins.
\newblock Catastrophic forgetting, rehearsal and pseudorehearsal.
\newblock \emph{Connection Science}, 1995.

\bibitem[Rostami(2021)]{rostami2021lifelong}
Mohammad Rostami.
\newblock Lifelong domain adaptation via consolidated internal distribution.
\newblock In \emph{NeurIPS}, 2021.

\bibitem[Saporta et~al.(2022)Saporta, Douillard, Vu, P{\'e}rez, and
  Cord]{saporta2022muhdi}
Antoine Saporta, Arthur Douillard, Tuan-Hung Vu, Patrick P{\'e}rez, and
  Matthieu Cord.
\newblock Multi-head distillation for continual unsupervised domain adaptation
  in semantic segmentation.
\newblock In \emph{Proceedings of the IEEE/CVF Conference on Computer Vision
  and Pattern Recognition (CVPR) Workshop}, 2022.

\bibitem[Shmelkov et~al.(2017)Shmelkov, Schmid, and
  Alahari]{shmelkov2017incremental}
Konstantin Shmelkov, Cordelia Schmid, and Karteek Alahari.
\newblock Incremental learning of object detectors without catastrophic
  forgetting.
\newblock In \emph{Proceedings of the IEEE international conference on computer
  vision}, pages 3400--3409, 2017.

\bibitem[Simon et~al.(2021)Simon, Koniusz, and Harandi]{simon2021learning}
Christian Simon, Piotr Koniusz, and Mehrtash Harandi.
\newblock On learning the geodesic path for incremental learning.
\newblock In \emph{Proceedings of the IEEE/CVF conference on Computer Vision
  and Pattern Recognition}, pages 1591--1600, 2021.

\bibitem[Szab{\'o} et~al.(2021)Szab{\'o}, Jamali-Rad, and
  Mannava]{szabo2021tilted}
Attila Szab{\'o}, Hadi Jamali-Rad, and Siva-Datta Mannava.
\newblock Tilted cross-entropy (tce): Promoting fairness in semantic
  segmentation.
\newblock In \emph{Proceedings of the IEEE/CVF Conference on Computer Vision
  and Pattern Recognition}, pages 2305--2310, 2021.

\bibitem[Thrun(1998)]{thrun1998lifelonglearning}
Sebastian Thrun.
\newblock Lifelong learning algorithms.
\newblock In \emph{Springer Learning to Learn}, 1998.

\bibitem[Truong and Luu(2025)]{truong2025cross}
Thanh-Dat Truong and Khoa Luu.
\newblock Cross-view action recognition understanding from exocentric to
  egocentric perspective.
\newblock \emph{Neurocomputing}, 614:\penalty0 128731, 2025.

\bibitem[Truong et~al.(2020)Truong, Duong, Luu, Tran, and Le]{truong2020domain}
Thanh-Dat Truong, Chi~Nhan Duong, Khoa Luu, Minh-Triet Tran, and Ngan Le.
\newblock Domain generalization via universal non-volume preserving approach.
\newblock In \emph{2020 17th Conference on Computer and Robot Vision (CRV)},
  pages 93--100. IEEE, 2020.

\bibitem[Truong et~al.(2021)Truong, Duong, Le, Phung, Rainwater, and
  Luu]{truong2021bimal}
Thanh-Dat Truong, Chi~Nhan Duong, Ngan Le, Son~Lam Phung, Chase Rainwater, and
  Khoa Luu.
\newblock Bimal: Bijective maximum likelihood approach to domain adaptation in
  semantic scene segmentation.
\newblock In \emph{International Conference on Computer Vision}, 2021.

\bibitem[Truong et~al.(2022)Truong, Chappa, Nguyen, Le, Dowling, and
  Luu]{truong2022otadapt}
Thanh-Dat Truong, Ravi Teja~Nvs Chappa, Xuan-Bac Nguyen, Ngan Le, Ashley~PG
  Dowling, and Khoa Luu.
\newblock Otadapt: Optimal transport-based approach for unsupervised domain
  adaptation.
\newblock In \emph{2022 26th international conference on pattern recognition
  (ICPR)}, pages 2850--2856. IEEE, 2022.

\bibitem[Truong et~al.(2023{\natexlab{a}})Truong, Le, Raj, Cothren, and
  Luu]{Truong:CVPR:2023FREDOM}
Thanh-Dat Truong, Ngan Le, Bhiksha Raj, Jackson Cothren, and Khoa Luu.
\newblock Fredom: Fairness domain adaptation approach to semantic scene
  understanding.
\newblock In \emph{IEEE Conference on Computer Vision and Pattern Recognition
  (CVPR)}, 2023{\natexlab{a}}.

\bibitem[Truong et~al.(2023{\natexlab{b}})Truong, Nguyen, Raj, and
  Luu]{truong2023fairness}
Thanh-Dat Truong, Hoang-Quan Nguyen, Bhiksha Raj, and Khoa Luu.
\newblock Fairness continual learning approach to semantic scene understanding
  in open-world environments.
\newblock In \emph{NeurIPS}, 2023{\natexlab{b}}.

\bibitem[Truong et~al.(2024{\natexlab{a}})Truong, Helton, Moustafa, Cothren,
  and Luu]{truong2024conda}
Thanh-Dat Truong, Pierce Helton, Ahmed Moustafa, Jackson~David Cothren, and
  Khoa Luu.
\newblock Conda: Continual unsupervised domain adaptation learning in visual
  perception for self-driving cars.
\newblock In \emph{Proceedings of the IEEE/CVF Conference on Computer Vision
  and Pattern Recognition Workshops}, pages 5642--5650, 2024{\natexlab{a}}.

\bibitem[Truong et~al.(2024{\natexlab{b}})Truong, Prabhu, Wang, Raj, Gauch,
  Subbiah, and Luu]{truong2024eagle}
Thanh-Dat Truong, Utsav Prabhu, Dongyi Wang, Bhiksha Raj, Susan Gauch,
  Jeyamkondan Subbiah, and Khoa Luu.
\newblock {EAGLE}: Efficient adaptive geometry-based learning in cross-view
  understanding.
\newblock In \emph{The Thirty-eighth Annual Conference on Neural Information
  Processing Systems}, 2024{\natexlab{b}}.

\bibitem[Vaswani et~al.(2017)Vaswani, Shazeer, Parmar, Uszkoreit, Jones, Gomez,
  Kaiser, and Polosukhin]{vaswani2017attention}
Ashish Vaswani, Noam Shazeer, Niki Parmar, Jakob Uszkoreit, Llion Jones,
  Aidan~N Gomez, {\L}ukasz Kaiser, and Illia Polosukhin.
\newblock Attention is all you need.
\newblock \emph{Advances in neural information processing systems}, 30, 2017.

\bibitem[Volpi et~al.(2021)Volpi, Larlus, and Rogez]{volpi2021continual}
Riccardo Volpi, Diane Larlus, and Gr{\'e}gory Rogez.
\newblock Continual adaptation of visual representations via domain
  randomization and meta-learning.
\newblock In \emph{Proc. Conf. Comp. Vision Pattern Rec.}, 2021.

\bibitem[Wang et~al.(2024)Wang, Wu, and Qin]{wang2024incremental}
Huyong Wang, Huisi Wu, and Jing Qin.
\newblock Incremental nuclei segmentation from histopathological images via
  future-class awareness and compatibility-inspired distillation.
\newblock In \emph{Proceedings of the IEEE/CVF Conference on Computer Vision
  and Pattern Recognition}, pages 11408--11417, 2024.

\bibitem[Wang et~al.(2021)Wang, Zhang, Zang, Cao, Pang, Gong, Chen, Liu, Loy,
  and Lin]{wang2021seesaw}
Jiaqi Wang, Wenwei Zhang, Yuhang Zang, Yuhang Cao, Jiangmiao Pang, Tao Gong,
  Kai Chen, Ziwei Liu, Chen~Change Loy, and Dahua Lin.
\newblock Seesaw loss for long-tailed instance segmentation.
\newblock In \emph{Proceedings of the {IEEE} Conference on Computer Vision and
  Pattern Recognition}, 2021.

\bibitem[Xie et~al.(2021)Xie, Wang, Yu, Anandkumar, Alvarez, and
  Luo]{xie2021segformer}
Enze Xie, Wenhai Wang, Zhiding Yu, Anima Anandkumar, Jose~M. Alvarez, and Ping
  Luo.
\newblock Segformer: Simple and efficient design for semantic segmentation with
  transformers.
\newblock In \emph{NeurIPS}, 2021.

\bibitem[Xie et~al.(2025)Xie, Lu, Xiao, Wang, Zhang, and Liu]{xie2025early}
Zhengyuan Xie, Haiquan Lu, Jia-wen Xiao, Enguang Wang, Le Zhang, and Xialei
  Liu.
\newblock Early preparation pays off: New classifier pre-tuning for class
  incremental semantic segmentation.
\newblock In \emph{European Conference on Computer Vision}, pages 183--201.
  Springer, 2025.

\bibitem[Yang et~al.(2019)Yang, Zhan, Chen, Yan, Loy, and
  Lin]{yang2019learning}
Lei Yang, Xiaohang Zhan, Dapeng Chen, Junjie Yan, Chen~Change Loy, and Dahua
  Lin.
\newblock Learning to cluster faces on an affinity graph.
\newblock In \emph{Proceedings of the IEEE/CVF conference on computer vision
  and pattern recognition}, pages 2298--2306, 2019.

\bibitem[Yang et~al.(2020)Yang, Chen, Zhan, Zhao, Loy, and
  Lin]{yang2020learning}
Lei Yang, Dapeng Chen, Xiaohang Zhan, Rui Zhao, Chen~Change Loy, and Dahua Lin.
\newblock Learning to cluster faces via confidence and connectivity estimation.
\newblock In \emph{Proceedings of the IEEE/CVF conference on computer vision
  and pattern recognition}, pages 13369--13378, 2020.

\bibitem[Yang et~al.(2023)Yang, Li, Ling, Zhang, Wang, Huang, Ma, Hur, and
  Lin]{yang2023label}
Ze Yang, Ruibo Li, Evan Ling, Chi Zhang, Yiming Wang, Dezhao Huang, Keng~Teck
  Ma, Minhoe Hur, and Guosheng Lin.
\newblock Label-guided knowledge distillation for continual semantic
  segmentation on 2d images and 3d point clouds.
\newblock In \emph{Proceedings of the IEEE/CVF International Conference on
  Computer Vision}, pages 18601--18612, 2023.

\bibitem[Zhang and Gao(2024)]{zhang2024background}
Anqi Zhang and Guangyu Gao.
\newblock Background adaptation with residual modeling for exemplar-free
  class-incremental semantic segmentation.
\newblock \emph{arXiv preprint arXiv:2407.09838}, 2024.

\bibitem[Zhang et~al.(2022)Zhang, Xiao, Liu, Chen, and
  Cheng]{zhang2022representation}
Chang-Bin Zhang, Jia-Wen Xiao, Xialei Liu, Ying-Cong Chen, and Ming-Ming Cheng.
\newblock Representation compensation networks for continual semantic
  segmentation.
\newblock In \emph{Proceedings of the IEEE/CVF Conference on Computer Vision
  and Pattern Recognition}, pages 7053--7064, 2022.

\bibitem[Zhou et~al.(2017)Zhou, Zhao, Puig, Fidler, Barriuso, and
  Torralba]{ade20k_challenge}
Bolei Zhou, Hang Zhao, Xavier Puig, Sanja Fidler, Adela Barriuso, and Antonio
  Torralba.
\newblock Scene parsing through ade20k dataset.
\newblock In \emph{CVPR}, pages 633--641, 2017.

\bibitem[Zhu et~al.(2022)Zhu, Wang, Chen, Chen, and Jiang]{zhu2022balanced}
Jianggang Zhu, Zheng Wang, Jingjing Chen, Yi-Ping~Phoebe Chen, and Yu-Gang
  Jiang.
\newblock Balanced contrastive learning for long-tailed visual recognition.
\newblock In \emph{Proceedings of the IEEE/CVF Conference on Computer Vision
  and Pattern Recognition}, pages 6908--6917, 2022.

\end{thebibliography}
}

\clearpage

\section{Proof of Propositions 1 and 2}

\subsection{Proof of Proposition 1}

\noindent
\textbf{Proposition 1}: \textit{If the contrastive clustering loss $\mathcal{L}_{Cont}(;, \mathbf{c})$ achieve the optimal value, the enforcement $\ell_i$ between the feature and the cluster will converges to $\ell_i = L^{-1}$.}

\noindent
\textbf{\textit{Proof:}} Let us consider the optimization of the Eqn. (4) in the paper as follows:
\begin{equation}\label{eqn:min_loss_one_cluster}
\begin{split}
    &\min -\sum_{i=1}^L\log\frac{\exp(\mathbf{f}^t_{i} \times \mathbf{c})}{\sum_{\mathbf{f}'}\exp(\mathbf{f}' \times \mathbf{c})}  = -\sum_{i=1}^L\log \ell_i \\
    &\text{subject to} \quad \sum_{i=1}^{L}\ell_i = \ell
\end{split}
\end{equation}
where $\ell$ is the total enforcement between features $\mathbf{f}^t_i$ and cluster $\mathbf{c}$. Then, the optimization of Eqn. (4) in the paper can be rewritten by using Lagrange multiplier as follows:
\begin{equation}
    \mathcal{L}\left(\{\ell_{i}\}_{i=1}^L, \lambda\right) = -\sum_{i=1}^L\log\ell_i + \lambda(\sum_{i=1}^L\ell_i - \ell) 
\end{equation}
where $\lambda$ is the Lagrange multiplier. Then, the contrastive clustering loss in Eqn. (4) in the paper achieves minimum if and only if:
\begin{equation}
\begin{split}
    \frac{\partial \mathcal{L}\left(\{\ell_{i}\}_{i=1}^L, \lambda\right) }{\partial \ell_i} &= -\ell_i^{-1} + \lambda = 0 \\
    \frac{\partial \mathcal{L}\left(\{\ell_{i}\}_{i=1}^L, \lambda\right) }{\partial \lambda} &= \sum_{i=1}^L\ell_i - \ell = 0 \\ 
    \Rightarrow \mathcal{L}\left(\{\ell_{i}\}_{i=1}^L, \lambda\right) &= -L\log \frac{\ell}{L}
\end{split}
\end{equation}
As the total enforcement between features and the cluster is normalized, i.e., $\ell \in [0..1]$, the contrastive clustering loss $\mathcal{L}\left(\{\ell_{i}\}_{i=1}^L, \lambda\right)$ achieves minimum when $\log\ell = 0 \Rightarrow \ell = 1$. Then, the enforcement between a single feature and the cluster will be equal to $\ell_i = \frac{\ell}{L} = L^{-1}$.

\subsection{Proof of Proposition 2}

\noindent
\textbf{Proposition 2}: \textit{If the fairness contrastive clustering loss $\mathcal{L}^{\alpha}_{Cont}(;, \mathbf{c})$ achieve the optimal value, the enforcement $\ell_i$ between the feature and the cluster will converges to $\ell_i = (\alpha^{-1} + L)^{-1}$.}

\noindent
\textbf{\textit{Proof:}} We first define the the enforcement between transitive vector $\mathbf{v}$ and the cluster $\mathbf{c}$ as $\ell_{\mathbf{v}} = \frac{\exp(\mathbf{v} \times \mathbf{c})}{\sum_{\mathbf{f}'}\exp(\mathbf{f}' \times \mathbf{c})}$. Then, let us consider the optimization of Eqn. (5) in the paper as follows:
\begin{equation}\label{eqn:min_loss_one_cluster_alpha}
\begin{split}
    &\min -\sum_{i=1}^L\alpha\log\ell_i - \log\ell_{\mathbf{v}} \\
    &\text{subject to} \quad \sum_{i=1}^{L}\ell_i + \ell_{\mathbf{v}} = \ell
\end{split}
\end{equation}
Similar to Eqn. \eqref{eqn:min_loss_one_cluster}, Eqn. \eqref{eqn:min_loss_one_cluster_alpha} can be reformulated via Lagrange multiplier as follows:
\begin{equation}
    \mathcal{L}\left(\{\ell_{i}\}_{i=1}^L, \lambda\right) = -\sum_{i=1}^L\alpha\log\ell_i -\log\ell_{\mathbf{v}}+ \lambda(\sum_{i=1}^L\ell_i + \ell_{\mathbf{v}}- \ell) 
\end{equation}
Then, the fairness contrastive loss $\mathcal{L}^{\alpha}_{Cont}$ achieves minimum if and only if:
\begin{equation}\label{eqn:der_loss_alpha}
\begin{split}
    \frac{\partial \mathcal{L}\left(\{\ell_{i}\}_{i=1}^L, \lambda\right) }{\partial \ell_i} &= -\alpha\ell_i^{-1} + \lambda = 0 \\
    \frac{\partial \mathcal{L}\left(\{\ell_{i}\}_{i=1}^L, \lambda\right) }{\partial \ell_\mathbf{v}} &= -\ell_\mathbf{v}^{-1} + \lambda = 0 \\
    \frac{\partial \mathcal{L}\left(\{\ell_{i}\}_{i=1}^L, \lambda\right) }{\partial \lambda} &= \sum_{i=1}^L\ell_i + \ell_{\mathbf{v}} - \ell = 0 \\ 
    \Rightarrow \mathcal{L}\left(\{\ell_{i}\}_{i=1}^L, \lambda\right) &= -
    \alpha L\log \frac{\alpha\ell}{1+\alpha L} - \log \frac{\ell}{1+\alpha L}
\end{split}
\end{equation}
As in Eqn. \eqref{eqn:der_loss_alpha}, the fairness contrastive learning loss $\mathcal{L}\left(\{\ell_{i}\}_{i=1}^L, \lambda\right)$ archives minimum when $\log\ell = 0 \rightarrow \ell = 1$. Thus, the enforcement between the single feature the cluster will be re-balanced as $\ell_i = \frac{\alpha}{1+\alpha L} = (\alpha^{-1}+L)^{-1}$.

\section{Implementation}

\noindent
\textbf{Implementation}
Our framework is implemented in PyTorch and trained on four 40GB-VRAM NVIDIA A100 GPUs.
The contrastive loss in our implementation is normalized with respect to the number of samples.
These models are optimized by the SGD optimizer \cite{bottou2010large} with momentum 0.9, weight decay $10^{-4}$, and a batch size of $16$. 
The learning rate of the first learning step and the continual steps is set to $10^{-4}$ and $5\times10^{-5}$ respectively.
To update the cluster vectors $\mathbf{c}$, following prior work \cite{truong2023fairness, joseph2021towards, he2020momentum}, we maintain a set of $500$ features for each cluster and update the clusters after $K = 100$ steps with a momentum $\eta = 0.99$. 
In our domain incremental experiments, all clusters are updated at each learning step by momentum update.
The number of features selected for each cluster in the visual grammar model is set to $M = 128$.
The balanced weight of CSS objective $\lambda_{CL}$ and the cluster regularizer $\lambda_{C}$ is set to $1$. Following the common practices \cite{truong2023fairness, joseph2021towards}, the margin between clusters $\nabla$ is set to 10.

\noindent
\textbf{Unknown Cluster Initialization}
As mentioned in the main paper, we adopt the DB-SCAN algorithm to initialize the clusters for unknown samples. In addition, to reduce the noise clusters and isolated clusters, we also merge several close clusters, i.e., if the distance between two clusters is less than the margin $2\nabla$, these will be merged into a single cluster where the new cluster center will be the means of these two merging cluster centers.
By empirical observation, we have noticed that the number of unknown clusters initialized at each learning step, i.e., $N_U$ at the current learning step $t$, is not greater than 1.5$\times$ times of the remaining classes (i.e., $|\mathcal{C}^{t+1..T}|$) in the dataset, e.g., in our ADE20K 100-50 experiments, at the first learning step of $100$ classes, there are $68$ unknown clusters that have been initialized while there are $50$ remaining unknown classes in the dataset. 

\noindent
\textbf{Cluster Assignment} In our approach, we use our visual grammar model to assign the cluster for each feature representation.
Theoretically, although there is a possibility that a feature could not be assigned to a cluster via the visual grammar model, we have empirically observed that this issue rarely happens in our approach. 
Indeed, since we initialize the known clusters via the DB-SCAN, it guarantees that for each feature, there is at least one cluster nearby that the feature representation should belong to.
However, to preserve the integrity of our approach, for the outlier features in cases that cannot be assigned clusters via the visual grammar model, 
these outliers will be heuristically assigned to their closest clusters as similar to \cite{joseph2021towards, truong2023fairness}.

\noindent
\textbf{Continual Learning Procedure} Algorithm \ref{algo:css_training} illustrates the training procedure of our CSS approach.

\begin{algorithm}\small
\caption{CSS Procedure At Learning Step $t$}
\label{algo:css_training}
\begin{algorithmic}[1]
\Require{
Learning Step $t$, Dataset $\mathcal{D}^t$, Visual Grammar $\phi(; \Theta_{t-1})$, and Segmentation Model $F(; \Theta_{t-1})$
}

\State \textbf{Step 0:} Extract features on $\mathcal{D}^t$ by $F(;, \theta_{t-1})$

\State \textbf{Step 1:} Initialize new known clusters for $\mathcal{C}^t$ of features extracted in \textbf{Step 0}

\State \textbf{Step 2:} Initialize potential unknown clusters of features extracted in \textbf{Step 0}

\State \textbf{Step 3:} Train CSS Model $F(; \theta_t)$ on $\mathcal{D}^t$

\State \textbf{Step 4:} Extract features on $\mathcal{D}^t$ by $F(;, \theta_{t})$

\State \textbf{Step 5:} Train Visual Grammar Model $\phi(;, \Theta_t)$ on current known clusters $\mathbf{c}$ and features extracted in \textbf{Step 4}

\State \Return $F(; \theta_t)$ and $\phi(; \Theta_t)$
\end{algorithmic}
\end{algorithm}

\section{Additional Experimental Results}

\subsection{Experiment Results of ADE20K 50-50 Benchmark}

Table \ref{tab:ade20k_50_50} presents the results of our method on the ADE20K 50-50 benchmark compared to prior methods. For fair comparisons, we use the DeepLab-V3 and Transformer in this experiment. As shown in the results, our proposed FALCON approach significantly outperforms prior methods. The results of our approach have reduced the gap with the upper bound result.

\begin{table}[H]
\centering
\setlength{\tabcolsep}{3pt}
\caption{Experimental results on ADE20K 50-50 Benchmark} \label{tab:ade20k_50_50}
\begin{tabular}{l | l|ccc}
\hline
\multicolumn{5}{c}{ADE20K 50-50 (3 steps)}\\
\hline
Network & Method    & 0-50 & 50-150 & all  \\
& MiB \cite{cermelli2020modelingthebackground}       & 45.6 & 21.0   & 29.3 \\
& PLOP \cite{douillard2021plop}      & 48.8 & 21.0   & 30.4 \\
& LGKD+PLOP \cite{yang2023label} & 49.4 & 29.4   & 36.0 \\
DeepLab-V3 & RCIL \cite{zhang2022representation}      & 47.8 & 23.0   & 31.2 \\
& RCIL+LGKD \cite{yang2023label} & 49.1 & 27.2   & 34.4 \\
& FairCL \cite{truong2023fairness}    & 49.7 & 26.8   & 34.6 \\
& \textbf{FALCON} & \textbf{50.6} & 	\textbf{31.2} &	\textbf{37.6} \\
\cdashline{2-5}
& Upper Bound & 51.1 &  33.25 & 38.9 \\
\hline
 & FairCL \cite{truong2023fairness} & 49.6 & 27.8 & 35.6 \\
Transformer & \textbf{FALCON}  & \textbf{53.0}	& \textbf{36.8} &	\textbf{42.2} \\
\cdashline{2-5}
  & Upper Bound  & 54.9 & 40.8 & 45.5 \\ 
\hline
\end{tabular}
\end{table}

\subsection{Ablation Study}

\noindent
\textbf{Effectiveness of Choosing Margin $\nabla$}
Table \ref{tab:abl_margin} studies the effectiveness of the value of margin $\nabla$ to the performance of our approach on ADE20K 100-50 and ADE20K 100-10 benchmarks. As shown in the results, the change of $\nabla$ also slightly influences the performance of the model. Since the margin defines the distance between two clusters, while the smaller value of the margin $\nabla$ could cause the incorrect cluster assignment of the features, the larger value of the margin $\nabla$ could produce the less compact clusters.

\begin{table}[H]
\centering
\caption{Effectiveness of Choosing Margin $\nabla$} \label{tab:abl_margin}
\begin{tabular}{l| c c c c c }
\hline
\multicolumn{6}{c}{(a) ADE20K 100-50}                    \\
\hline
                & 0-100 & 101-150 & all  & Major & Minor \\
\hline
$\nabla=5$     &  44.4 & 21.8 & 36.9 & 51.9 & 29.4 \\
$\nabla=10$     & \textbf{44.6} & \textbf{24.5} & \textbf{37.9} & \textbf{52.1} & \textbf{30.8} \\
$\nabla=20$     &  44.7 & 22.2 & 37.2 & 51.7 & 29.9  \\
\hline
\multicolumn{6}{c}{(b) ADE20K 100-10}                    \\
\hline

                & 0-100 & 101-150 & all  & Major & Minor \\
\hline
$\nabla=5$     &  43.2 & 18.7 & 35.0 & 50.5 & 27.3 \\
$\nabla=10$     &  \textbf{44.4}  & \textbf{20.4}    & \textbf{36.4} & \textbf{51.8}  & \textbf{28.7}  \\
$\nabla=20$     &  43.5 & 19.9 & 35.7 & 51.2 & 27.9 \\
\hline
\end{tabular}
\end{table}

\noindent
\textbf{Effectiveness of Choosing Number of Features $M$}
We study the impact of choosing the number of features $M$ in the visual grammar model. As in shown Table \ref{tab:abl_visual_grammar_m}, the optimal performance of our approach is $M = 128$.
When the number of features selected is small ($M = 96$), it does not have enough number of features to form the visual grammar so the model is hard to exploit the correlation among features and the cluster. 
Meanwhile, when we increase the number of selected features ($M = 256$), the clusters will consist of many outlier features (the ones that do not belong to the cluster), thus being challenging for the visual grammar model to exploit the topological structures of the feature distribution.

\begin{table}[H]
\centering
\caption{Effectiveness of Number of Features $M$ in a Cluster of Visual Grammar Model.} \label{tab:abl_visual_grammar_m}
\begin{tabular}{l| c c c c c }
\hline
\multicolumn{6}{c}{(a) ADE20K 100-50}                    \\
\hline
                & 0-100 & 101-150 & all  & Major & Minor \\
\hline
$M=96$     &  43.0 &	19.6 &	35.2 & 50.5 & 27.5  \\
$M=128$     & \textbf{44.6} & \textbf{24.5} & \textbf{37.9} & \textbf{52.1} & \textbf{30.8} \\
$M=256$     &  43.6 &	21.6 &	36.3 & 51.0 & 28.9 \\ 
\hline
\multicolumn{6}{c}{(b) ADE20K 100-10}                    \\
\hline

                & 0-100 & 101-150 & all  & Major & Minor \\
\hline
$M=96$     &  42.2 &	16.4 &	33.6 & 50.2 & 25.3 \\
$M=128$     &  \textbf{44.4}  & \textbf{20.4}    & \textbf{36.4} & \textbf{51.8}  & \textbf{28.7}  \\
$M=256$     &  42.7 &	17.1 &	34.2 & 50.6 & 26.0 \\
\hline
\end{tabular}
\end{table}

\noindent
\textbf{Effectiveness of Different Segmentation Networks}
To illustrate the flexibility of our proposed approach, we evaluate our proposed approach with different network backbones. Table \ref{tab:abl_different_backbones} illustrates the results of our approach using DeepLab-V3 \cite{chen2018deeplab}, SegFormer \cite{xie2021segformer} with different backbones, i.e., ResNet-50, ResNet-101, MiT-B2, and MiT-B3.
As shown in the performance, the more powerful the segmentation model is, the better performance of the model is.
In particular, our approach has shown its flexibility since it consistently improves the performance of the segmentation model and achieves the SOTA performance on two different benchmarks, i.e., the performance of Transformer models achieves $41.9\%$, and $40.3\%$ on ADE20K 100-50, ADE20K 100-10, respectively.

\begin{table}[H]
\small
\centering
\caption{Effectiveness of Different Backbones on ADE20K.} \label{tab:abl_different_backbones}
\setlength{\tabcolsep}{3pt}
\begin{tabular}{l | c | c c c c c }
\hline
\multicolumn{7}{c}{(a) ADE20K 100-50}                    \\
\hline
   & Backbone             & 0-100 & 101-150 & all  & Major & Minor \\
\hline

\multirow{2}{*}{DeepLab-V3}  &R-50       &  44.3 & 15.2 & 34.7   & 51.5 & 26.4    \\
 & R-101     & \textbf{44.6} & \textbf{24.5} & \textbf{37.9} & \textbf{52.1} & \textbf{30.8} \\
 \hline
\multirow{2}{*}{Transformer} & MiT-B2       & 44.5 & 27.4 & 38.8  & 52.4 & 32.2      \\
 & MiT-B3      & \textbf{47.5} & \textbf{30.6} & \textbf{41.9}  & \textbf{53.8} & \textbf{35.8}   \\

\hline
\multicolumn{7}{c}{(b) ADE20K 100-10}                    \\
\hline
    
    & Backbone            & 0-100 & 101-150 & all  & Major & Minor \\
\hline

\multirow{2}{*}{DeepLab-V3} & R-50       &  43.5	& 16.5 &	34.5 & 51.1 & 26.2     \\
 & R-101     &  \textbf{44.4}  & \textbf{20.4}    & \textbf{36.4} & \textbf{51.8}  & \textbf{28.7}  
\\
\hline
\multirow{2}{*}{Transformer} & MiT-B2       &  45.4 &	22.7 &	37.8 & 52.6 & 30.4    \\
 & MiT-B3       & \textbf{47.3} & \textbf{26.2} & \textbf{40.3} & \textbf{54.0} & \textbf{33.4}   \\
\hline
\end{tabular}
\end{table}

\section{Relation to Knowledge Distillation}

Knowledge Distillation is a common approach to continual semantic segmentation \cite{douillard2021plop, cermelli2023comformer,  zhang2022representation, ssul_neurips_2021}. Prior work in clustering \cite{truong2023fairness} has shown that the clustering loss is an upper bound of the knowledge distillation loss. 
Formally, the knowledge distillation loss can be formed as follows:
\begin{equation}
\small
    \mathcal{L}_{distill}(\mathbf{x}^t, F, \theta_t, \theta_{t-1}) = \mathcal{L}(\mathbf{F}^{t-1}, \mathbf{F}^t)
\end{equation}
where $\mathbf{F}^{t}$ and $\mathbf{F}^{t-1}$ are the feature representations extracted from the model at learning step $t$ and step $t-1$, respectively, and the metric $\mathcal{L}$ measure the knowledge gap between $\mathbf{F}^{t}$ and $\mathbf{F}^{t-1}$. Then, given a set of cluster $\mathbf{c}$, we consider the following triangle inequality of the metric $\mathcal{L}$ as follows:
\begin{equation} 
\small
\label{eqn:proof_ineq}
\begin{split}
    \forall \mathbf{c}: \quad \mathcal{L}(\mathbf{F}^{t}, \mathbf{F}^{t-1}) &\leq \mathcal{L}(\mathbf{F}^{t}, \mathbf{c}) + \mathcal{L}(\mathbf{c}, \mathbf{F}^{t-1}) \\
    \Leftrightarrow \underbrace{\mathcal{L}(\mathbf{F}^{t}, \mathbf{F}^{t-1})}_{\mathcal{L}_{distill}} &\leq \frac{1}{|\mathcal{C}^{1..T}|}\sum_{\mathbf{c}}\left[\underbrace{\mathcal{L}(\mathbf{F}^{t}, \mathbf{c})}_{\mathcal{L}_{Cont}} + \mathcal{L}(\mathbf{c}, \mathbf{F}^{t-1})\right]
\end{split}
\end{equation}
At the computational time of Contrastive Clustering loss, the set of cluster vectors $\mathbf{c}$ is fixed (could be considered as constants).
In addition, the features extracted at learning step $t-1$, i.e., $\mathbf{F}^{t-1}$, are constant due to the fix pre-trained model $\theta_{t-1}$. Therefore, without a strict argument, the distance $\mathcal{L}(\mathbf{c}, \mathbf{F}^{t-1})$ could be considered as constant. Therefore, Eqn. \eqref{eqn:proof_ineq} can be further derived as follows:
\begin{equation} \label{eqn:upper_bound_loss}
\footnotesize
\begin{split}
        \underbrace{\mathcal{L}(\mathbf{F}^{t}, \mathbf{F}^{t-1})}_{\mathcal{L}_{distill}} &= \mathcal{O}\Bigg(\underbrace{\mathcal{L}\frac{1}{|\mathcal{C}^{1..T}|}}_{Constant}\sum_{\mathbf{c}}\Big[\underbrace{\mathcal{L}(\mathbf{F}^{t}, \mathbf{c})}_{\mathcal{L}_{Cont}} + \underbrace{\mathcal{L}(\mathbf{c}, \mathbf{F}^{t-1})}_{\text{Constant}}\Big]\Bigg) \\
        &= \mathcal{O}\Bigg(\underbrace{\sum_{\mathbf{c}}\mathcal{L}(\mathbf{F}^{t}, \mathbf{c})}_{\mathcal{L}_{Cont}}\Bigg) \\
    \Rightarrow \mathcal{L}_{distill}(\mathbf{F}^{t-1}, \mathbf{F}^t)  &= \mathcal{O}\left(\mathcal{L}_{Cont}(\mathbf{F}^{t}, \mathbf{c})\right)
\end{split}
\end{equation}
where $\mathcal{O}$ is the Big-O notation. Hence, from Eqn. \eqref{eqn:upper_bound_loss}, without lack of generality, we can observe that the Contrastive Clustering Loss is the upper bound of the Knowledge Distillation loss.
Therefore, by minimizing the Contrastive Clustering Loss, the constraint of  Knowledge Distillation is also maintained due to the property of the upper bound.

\section{Discussion of Limitations and Broader Impact}

\textbf{Limitations.} In our paper,  we choose a specific set of hyper-parameters and learning approaches to support our hypothesis. However, our work could contain several limitations. First, choosing the scaling factor $\alpha$ could be considered as one of the potential limitations of our approach. In practice, when data keeps continuously growing, the pre-defined scaling factor $\alpha$ could not be good enough to control the fairness among classes.
Our work focuses on investigating the effectiveness of our proposed losses to fairness, catastrophic forgetting, and background shift problems. Thus, the investigation of balance weights among losses has not been fully exploited, and we leave this experiment as our future work.
Third, initializing the unknown clusters at each training step could potentially be room for improvement since the bad initial clusters could result in difficulty during training and updating these clusters and linking the unknown clusters learned in previous steps and new initial unknown clusters at the current learning steps have been yet fully exploited in our method. 
In addition, while our approach is designed for the DeepLab-V3 and Transformer segmentation networks \cite{chen2018deeplab, xie2021segformer}, the extensions of FALCON to mask-based segmentation networks \cite{cheng2022masked,cheng2021maskformer, cermelli2023comformer} could be a potential next research for further performance improvement.
These limitations could motivate new studies to further improve Fairness Learning via the Contrastive Attention Approach to continual learning in the future.

\noindent
\textbf{Broader Impact.} Our paper investigates and addresses the fairness problem in continual learning. Our contribution is a step toward the fairness and transparency of continual semantic segmentation.
Our study highlights the significance of fairness in continual semantic segmentation learning and presents a novel approach to address fairness issues, enhancing the robustness and credibility of the segmentation model.

\end{document}